\documentclass{article}
\usepackage{natbib}
\setcitestyle{numbers,square,comma}
\usepackage{multirow}

\usepackage[preprint]{neurips_2025}
\usepackage{etoolbox}
\newtoggle{isSubmission}
\togglefalse{isSubmission}

\usepackage{times}
\usepackage{epsfig}
\usepackage{graphicx}
\usepackage{amsmath}
\usepackage{amssymb}

\usepackage{booktabs} 
\usepackage{diagbox}
\usepackage{multirow}
\usepackage{ifthen}
\usepackage{lineno}
\usepackage{ifthen}
\usepackage{cases}
\usepackage{footnote}
\usepackage{lipsum}
\makesavenoteenv{table}
\usepackage{array}
\usepackage{makecell} % 用于在单元格中换行
\usepackage{tabularx} % 用于自动调整列宽
\usepackage{bbding}       % \Checkmark,\XSolid,... (需要和pifont宏包共同使用)
\usepackage{fontawesome}  % \faCheck,\faTimes

\usepackage{hyperref}       % hyperlinks
\hypersetup{
    colorlinks,
    linkcolor={red!85!black},
    citecolor={blue!70!cyan},
    urlcolor={blue!90!black}
}

\usepackage[utf8]{inputenc} % allow utf-8 input
\usepackage[T1]{fontenc}    % use 8-bit T1 fonts
\usepackage{url}            % simple URL typesetting
\usepackage{booktabs}       % professional-quality tables
\usepackage{amsfonts}       % blackboard math symbols
\usepackage{nicefrac}       % compact symbols for 1/2, etc.
\usepackage{microtype}      % microtypography

\usepackage{xspace}

\newcommand{\eg}{\textit{e.g.}\xspace}

\newcommand{\etc}{\textit{etc.}\xspace}

\newcommand*{\affaddr}[1]{#1} 
\newcommand*{\affmark}[1][*]{\textsuperscript{#1}}
\newcommand{\authormark}[2][]{%
  \begingroup
  \def\@thefnmark{#1}%
  \footnote{#2}%
  \endgroup
}

\title{VCapsBench: A Large-scale Fine-grained Benchmark for Video Caption Quality Evaluation}

\author{%
\bf
Shi-Xue Zhang \affmark[1,2]\thanks{Equal contribution}~~
Hongfa Wang\affmark[2,3]\textsuperscript{*}~~
Duojun Huang\affmark[2]\textsuperscript{*}~~ 
Xin Li\affmark[2]\textsuperscript{*}\\
\bf
Xiaobin Zhu\affmark[1]\textsuperscript{$^\dagger$}~~ 
Xu-Cheng Yin\affmark[1]\thanks{Corresponding author}~~~ \\
\affaddr{\affmark[1]University of Science and Technology Beijing}
\affaddr{\affmark[2]Tencent} \\
\affaddr{\affmark[3]Tsinghua Shenzhen
International Graduate School}
}

\usepackage{graphics}
\usepackage[table]{xcolor}
\usepackage{subcaption}
\usepackage{caption}
\usepackage{amssymb}
\usepackage{algorithm}
\usepackage{algpseudocode}
\usepackage{wrapfig}
\usepackage{longtable}
\usepackage{fancyvrb}
\usepackage{empheq}
\definecolor{metacolor}{HTML}{0064E0}
\hypersetup{
  colorlinks,
  linkcolor=metacolor,
  citecolor=metacolor,
  urlcolor=metacolor
}
\usepackage{adjustbox}
\usepackage{makecell} 
\definecolor{color_blue}{HTML}{E7EFFA}
\definecolor{color_green}{HTML}{E6F8E0}
\definecolor{color_gray}{HTML}{ECECEC}
\definecolor{pearDark}{HTML}{2980B9}

\begin{document}

\maketitle
\begin{abstract}
Video captions play a crucial role in text-to-video generation tasks, as their quality directly influences the semantic coherence and visual fidelity of the generated videos. Although large vision-language models (VLMs) have demonstrated significant potential in caption generation, existing benchmarks inadequately address fine-grained evaluation, particularly in capturing spatial-temporal details critical for video generation. To address this gap, we introduce the Fine-grained Video Caption Evaluation Benchmark (\textbf{VCapsBench}), the first large-scale fine-grained benchmark comprising 5,677 (5K+) videos and 109,796 (100K+) question-answer pairs. These QA-pairs are systematically annotated across 21 fine-grained dimensions (\eg, camera movement, and shot type) that are empirically proven critical for text-to-video generation. We further introduce three metrics (Accuracy (\textbf{AR}), Inconsistency Rate (\textbf{IR}), Coverage Rate (\textbf{CR})), and an automated evaluation pipeline leveraging large language model (LLM) to verify caption quality via contrastive QA-pairs analysis. By providing actionable insights for caption optimization, our benchmark can advance the development of robust text-to-video models. The dataset and codes are available at website: \url{https://github.com/GXYM/VCapsBench}. 
\end{abstract}

\begin{figure}[ht]
	\begin{minipage}[t]{1.0\linewidth}
        \centering
	\includegraphics[width=0.8\linewidth]{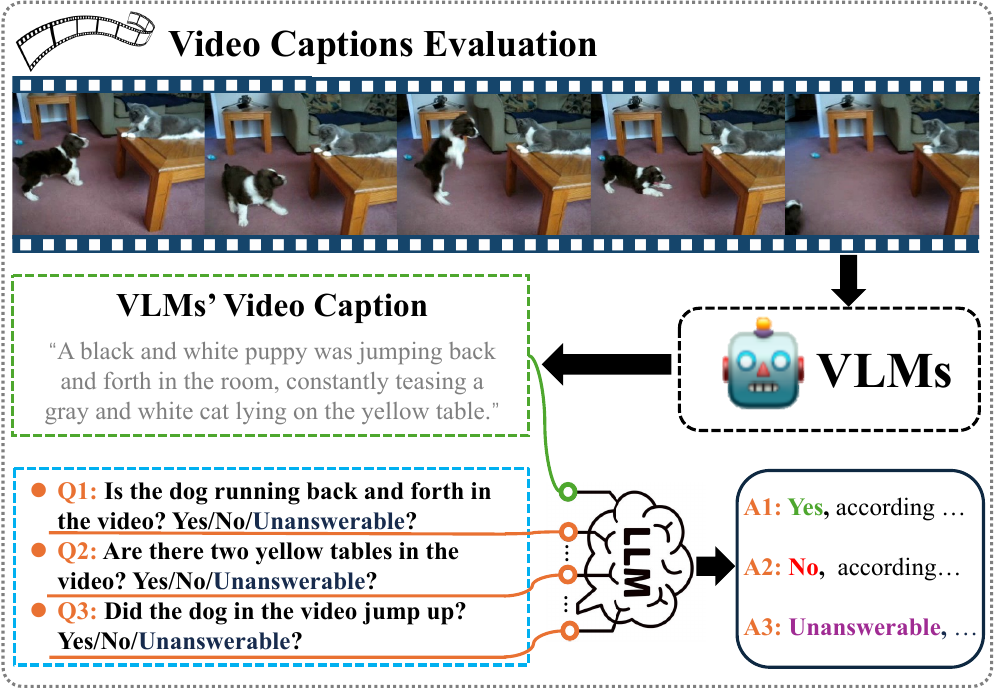}
	\caption{Illustration of video caption evaluation by VCapsBench. Evaluate the detail, comprehensiveness, and accuracy of video captions using "yes-no" question-answer pairs.}
	\label{fig:fig1}
	\end{minipage}%
	\vspace{-1.5em}	
\end{figure}    
\section{Introduction}
\label{sec:intro}
Recent advances in video understanding~\cite{reid2024gemini, liu2025st, hong2024cogvlm2, liu2024oryx, zhang2024STGT, li2024aria} and generation~\cite{SORA, vidu, kelin, hunyuanvideo} have been driven by large vision-language models (VLMs). For video comprehension, researchers have extended image-based architectures (\eg, LLaVA-Video~\cite{zhang2024video}, PLLaVA~\cite{xu2024pllava} and CogVLM2-Video~\cite{hong2024cogvlm2}) and explored hybrid image-video training approaches (Qwen2-VL~\cite{wang2024qwen2}, LLaVA-OneVision~\cite{li2024llava}). Currently, video generation systems such as Sora~\cite{SORA} and HunyuanVideo~\cite{hunyuanvideo} use VLMs for multimodal captioning and prompt engineering. However, current benchmarks~\cite{li2024mvbench, li2024videovista, wang2024lvbench, fu2024video, zhou2024mlvu, wu2024longvideobench} struggle to assess the detailed spatio-temporal aspects required for these applications.

In visual generation, existing caption evaluation metrics fall mainly into two categories: reference-based (METEOR~\cite{meteor}, BLEU~\cite{bleu}, SPICE~\cite{Spice}, and CIDEr~\cite{cider}) and reference-free (InfoMetIC~\cite{infometic}, CLIPScore~\cite{Clipscore}, and TIGEr~\cite{Tiger}). Reference-based metrics~\cite{meteor,bleu,Spice,cider} assess the quality of the captions by comparing them with ground-truth captions. However, these scores are highly dependent on the reference captions' format. Hence, FAIEr~\cite{Faier} adopts visual and textual scene graphs for a more robust reference-based evaluation. Reference-free metrics~\cite{infometic,Clipscore,Tiger} use semantic vectors from the reference image to evaluate the similarity of the caption. These methods falter with concept-dense captions, overwhelmed by numerous concepts. Recent innovations like QACE~\cite{QACE}, DSG ~\cite{cho2023davidsonian}], DPG-bench~\cite{Ella} and CapsBench~\cite{playgroundv3} employ question-answering frameworks to address dense concept evaluation for image captioning. Notably, these methods derived from image description evaluation still have significant measurement blind spots when dealing with the spatio-temporal dynamic elements of video generation scenarios, \eg,  camera motion (``slow zoom" vs. ``fast pan") or dynamic spatial relationships (object displacement from ``left foreground " to ``central midground"). 
In these cases, video generation systems (\eg, Sora) need to ensure accurate text-video alignment.

Current video understanding benchmarks primarily focus on holistic semantic alignment (MVBench ~\cite{li2024mvbench}, VideoVista~\cite{li2024videovista}) or specific skill evaluation (LVBench~\cite{wang2024lvbench}), neglecting fine-grained spatial-temporal dynamics essential for video generation. This gap becomes critical when evaluating text-to-video systems where caption precision directly impacts visual output. For instance, failing to distinguish between ``slow zoom" versus ``fast pan" camera motions, or misrepresenting object trajectories from ``left foreground" to ``central midground". While manual evaluation of such nuances remains impractical at scale, automated metrics also struggle with dynamic semantic alignment due to inherent limitations in traditional evaluation paradigms. To address this problem, it is necessary to re-examine the design principles of video description evaluation metrics.

To address critical gaps in evaluating video caption,  we introduce \textbf{VCapsBench}, the first large-scale fine-grained benchmark for video caption evaluation. It contains 5,677 diverse videos with 109,796 human-verified questions, allowing fine-grained evaluation of caption quality. Specifically, we employ text-based question-answering across 21 categories, including action, camera movements (\eg, zoom, pan and tilt), object positioning (absolute or relative position), entity and shot type, \etc, with ``yes", ``no", and ``unanswerable" ternary judgments. Unlike image-focused caption benchmarks like CapsBench~\cite{playgroundv3}, our VCapsBench prioritizes temporal continuity through video-specific queries (avg. 19 questions/video) while mitigating LLM hallucination via an ``Unanswerable" option. To accurately and comprehensively assess caption quality, we introduce three metrics: Accuracy (AR), Inconsistency Rate (IR), and Coverage Rate (CR). To automate the calculation of these metrics, we develop an evaluation pipeline that leverages powerful LLMs for objective video caption quality measurement, as illustrated in Fig.~\ref{fig:fig1}. In summary, our contributions are as follows:
\begin{itemize}
    \item We introduce \textbf{VCapsBench}, the first large-scale fine-grained benchmark for video caption evaluation, featuring diverse videos (5K+) and QA-pairs (100K+).
    \item We introduce three metrics: Accuracy (\textbf{AR}), Inconsistency Rate (\textbf{IR}), and Coverage Rate (\textbf{CR}), along with an automated evaluation pipeline to fairly assess the correctness and coverage of video captions.
    
    \item We evaluate ten VLMs, including seven advanced open models (Qwen2VL, Qwen2.5VL, InternVL2.5, LLaVA-Video, and VideoLLaMA3, \etc) and three closed model, GPT-4o,  Gemini2.5-Pro Flash and Preview, providing a solid reference for the community.
\end{itemize}

\section{Related Work}
\label{sec:RelatedWork}

\textbf{Reference-based metric methods}~\cite{meteor, bleu, Spice, cider, Faier} evaluate caption quality by comparing generated captions with ground-truth captions. BLEU~\cite{bleu} is a fast, cost-effective, and language-independent metric for machine translation, correlating well with human assessments. METEOR~\cite{meteor} also evaluates machine translation and shows a high correlation with human judgments, significantly outperforming BLEU. CIDEr~\cite{cider} measures the similarity of generated sentences against a set of human-written ground-truth sentences, serving as an automatic consensus metric for image description quality. However, these metrics primarily focus on n-gram overlap, which is neither necessary nor sufficient for simulating human judgment. To address these limitations, SPICE introduces a metric based on semantic propositional content over scene graphs, though its scores highly rely on the format of reference captions.  Newer methods like FAIEr~\cite{Faier} leverage visual and textual scene graphs for a more robust evaluation.

\textbf{Reference-free metric methods}~\cite{infometic, Clipscore, Tiger} utilize semantic vectors from the reference image to assess caption similarity. InfoMetIC~\cite{infometic} is an informative metric for reference-free image caption evaluation, capable of identifying incorrect words and unmentioned image regions with fine-grained precision. It provides a text precision score, a vision recall score, and an overall quality score at a coarse-grained level, with the latter showing significantly better correlation with human judgments than existing metrics across multiple benchmarks. CLIPScore~\cite{Clipscore} employs CLIP~\cite{clip} for robust automatic evaluation of image captioning without references, focusing on image-text compatibility. Although complementary to reference-based metrics that emphasize text-text similarities, CLIPScore is relatively weaker for tasks requiring richer contextual knowledge, such as news captions. TIGEr~\cite{Tiger} assesses caption quality by evaluating both the representation of image content and the alignment of machine-generated captions with human-generated ones.

\textbf{Question-based evaluation methods}~\cite{QACE, cho2023davidsonian, playgroundv3} have been developed to enhance the assessment of caption quality in visual generation tasks. The QACE framework~\cite{QACE} generates questions from captions to evaluate their quality. A similar approach has been proposed for image generation models, where the Davidsonian Scene Graph (DSG)~\cite{cho2023davidsonian} organizes questions into dependency graphs, facilitating comprehensive evaluation of text-to-image models. Inspired by DSG and DPG-bench~\cite{Ella}, Playground v3~\cite{playgroundv3} introduces CapsBench, a benchmark for image captioning that uses ``yes-no" question-answer pairs. However, these benchmarks focus solely on image captioning, while video captioning requires consideration of additional factors such as actions, motion, camera movements, and shot types, which are crucial for text-to-video generation.
\section{The VCapsBench Benchmark}

\subsection{Overview}
Although existing benchmarks can assess the quality of VLM-generated image captions across various dimensions, as discussed in Sec.~\ref{sec:RelatedWork}, they still fall short in evaluating the fine-grained video captions produced by VLMs. To effectively measure the understanding and descriptive capabilities of VLMs across different types of video, we have constructed a large-scale high-quality benchmark VCapsBench for video caption quality evaluation.

\begin{figure*}[tp]
	\begin{minipage}[t]{1.0\linewidth}
        \centering
	\includegraphics[width=0.95\linewidth]{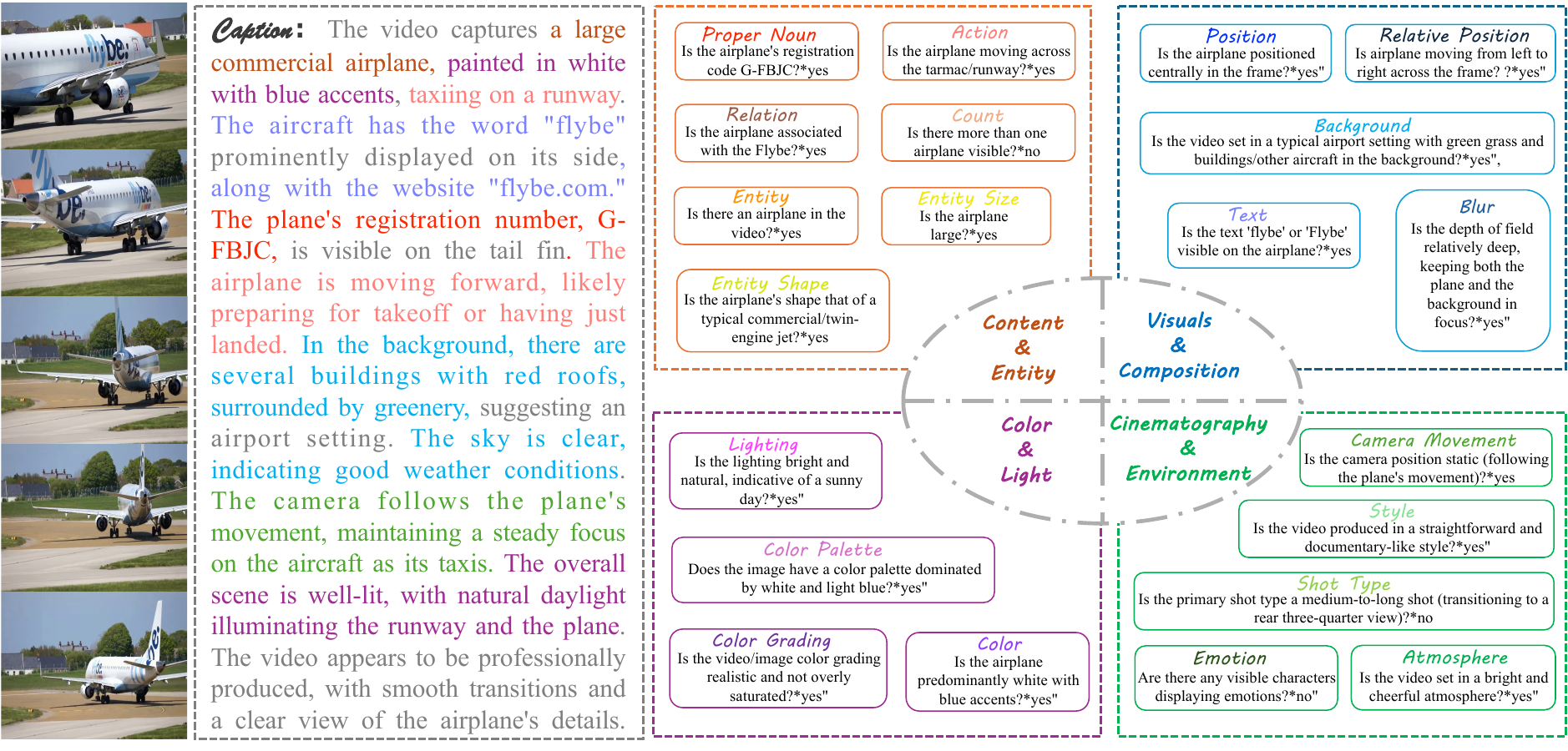}
	\caption{An example of video caption and  question-answer pairs in our VCapsBench.}
	\label{fig:fig2}
	\end{minipage}%
	\vspace{-1.0em}	
\end{figure*}

\begin{figure*}[tp]
	\begin{minipage}[t]{1.0\linewidth}
        \centering
	\includegraphics[width=0.95\linewidth]{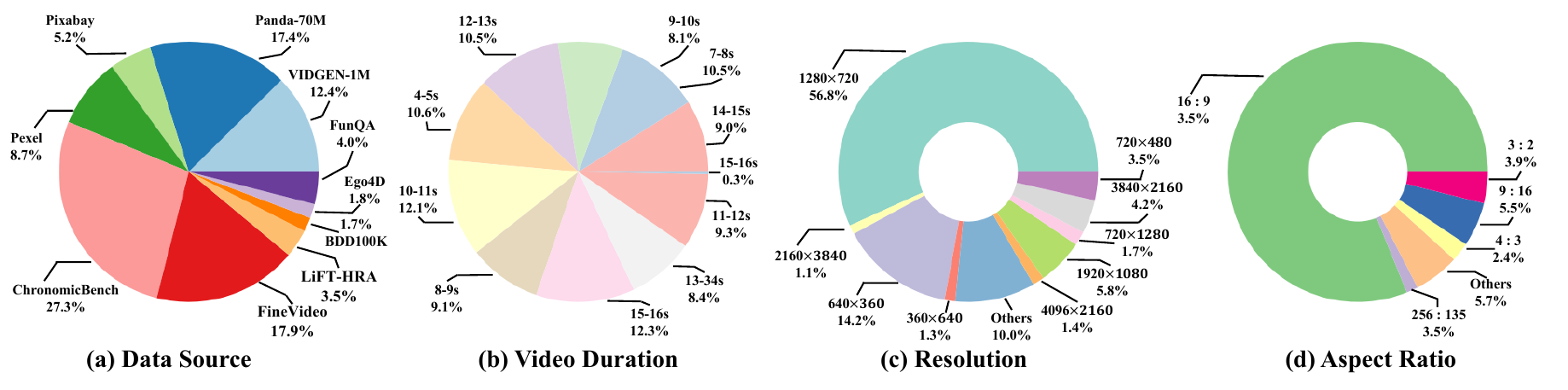}
	\caption{(a) Video source distribution; (b) Video duration distribution; (c) Video resolution distribution (d) Video aspect ratio distribution.}
	\label{fig:fig3}
	\end{minipage}%
	\vspace{-1.5em}	
\end{figure*}

\subsection{Dataset Statistics}
\textbf{Data Dimensions.} As shown in Fig.~\ref{fig:fig2}, the evaluation of video captions is organized into four primary categories: ``Content and Entity", ``Visuals and Composition" ``Color and Lighting" and ``Cinematography and Atmosphere". ``Content and Entity" includes seven subcategories, which focus on core elements like action, count, entity, entity size, entity shape, proper noun and relation. ``Visuals and Composition" is related to spatial arrangement and visual presentation, comprising five subcategories: position, relative position, text, blur, and background. ``Color and Lighting" addresses how these elements enhance mood and style, covering four subcategories: color, color palette, color grading, and lighting. Finally, ``Cinematography and Atmosphere" includes five subcategories: camera movement, shot type, style, atmosphere, and emotion, integrating filming techniques and artistic expression.  By categorizing these dimensions, it becomes easier to statistically analyze the quality of video  captions across different categories. This provides more detailed guidance for optimizing VLM's video understanding capabilities.

\textbf{Data Collection.} To fully support the caption evaluation task in video understanding and generation tasks, we prioritized  the complexity of aesthetic and content in our data collection for VCapsBench. We sourced videos from 10 publicly available datasets to ensure diversity: Panda-70M\cite{ChenSMDCJF0RYT24}, Ego4D\cite{grauman2022ego4d}, BDD100K\cite{yu2020bdd100k}, Pixabay~\cite{chensharegpt4video}, Pexel~\cite{chensharegpt4video}, VIDGEN-1M\cite{tan2024vidgen}, ChronomicBench~\cite{yuan2024chronomagic}, FineVideo~\cite{Farre2024FineVideo}, FunQA~\cite{yuan2024chronomagic}, and LiFT-HRA-20K~\cite{wang2024lift}. As illustrated in Fig.~\ref{fig:fig3}, we curated a diverse collection of 988 high-resolution videos from Panda-70M\cite{ChenSMDCJF0RYT24}, encompassing a wide range of scenes such as wildlife, cooking, sports, TV shows, gaming, and 3D rendering. These videos often contain complex content and transformations, providing a robust foundation for understanding various real-world scenarios. Additionally, we included 494 high-resolution videos from Pexels and 298 from Pixabay, both renowned for their scenic landscapes and human activities, characterized by high aesthetic quality and detailed imagery. To further ensure data diversity, we sample 1,018 videos from FineVideo\cite{Farre2024FineVideo}, which encompasses 6 major categories and 122 subcategories. Despite their lower resolution (below 640 $ \times $360), we balanced this by sampling 333 ultra-high-resolution videos (2K, 3K, and 4K) from VIDGEN-1M\cite{tan2024vidgen}, typically used for training text-to-video models due to their high detail and quality. Our collection was further enriched with videos from Ego4D\cite{grauman2022ego4d} and BDD100K\cite{yu2020bdd100k} to cover ego-centric human activities and autonomous driving scenarios, ensuring a comprehensive representation of real-world scenes.
To evaluate the VLM model's understanding of object motion and physical laws, we included 1,549 videos from ChronomicBench, spanning four main categories
(\eg, biological, artificial, meteorological, and physical) across 75 subcategories. To further diversify our dataset, we sampled 227 videos from FunQA and 200 from LiFT-HRA-20K. FunQA features human-centric content such as humorous clips, creative performances, and visual illusions, while LiFT-HRA-20K consists of synthetic video data.

As shown in Fig.~\ref{fig:fig3}, our VCapsBench dataset comprises 5,677 videos from a wide range of scenes, e.g., natural landscapes, animals, human activities, physical phenomena, games, 3D renderings, and synthetic videos (more than 100 subcategories). VCapsBench also features diverse video durations (4 to 16 seconds), resolutions (125 different resolutions), and aspect ratios (87 different ratios). This extensive collection allows for a thorough evaluation of VLM models' understanding and insight across various video types, ensuring a robust assessment of video captioning.

\begin{figure*}[thbp]
	\begin{minipage}[t]{1.0\linewidth}
        \centering
	\includegraphics[width=0.9\linewidth]{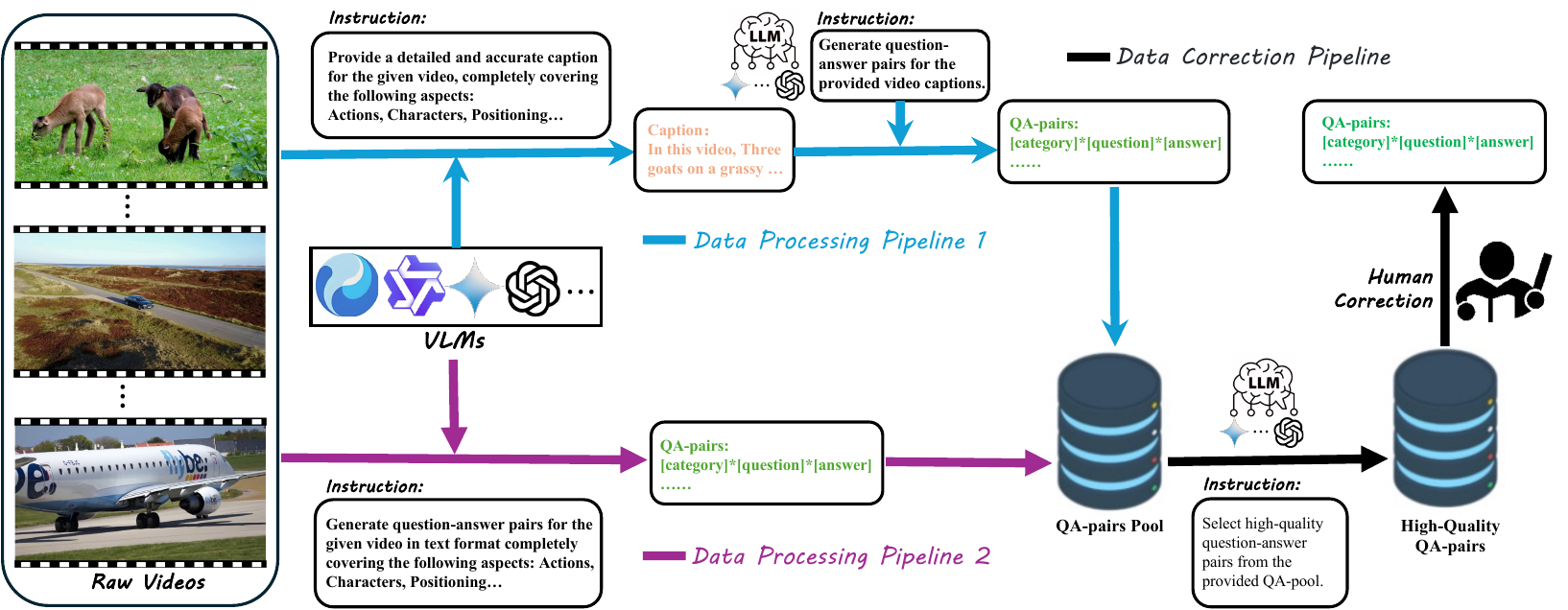}
	\caption{The pipeline of QA-pairs generation, which includes multiple data processing pipelines and a data correction pipeline.}
	\label{fig:fig4}
	\end{minipage}%
	\vspace{-1.0em}	
\end{figure*}

\subsection{Annotation Details}
Inspired by DPG-bench~\cite{Ella} and CapsBench~\cite{playgroundv3}, we developed a method to evaluate video caption quality using multi-dimensional question-answer pairs. Based on the raw video, we generate ``yes-no" question-answer pairs as annotations. As shown in Fig.~\ref{fig:fig4}, our ``Data Processing Pipeline" creates these pairs, which are stored in the ``QA-pairs Pool". Each pair includes a category, question, and answer, formatted as ``[category][question][answer]" for easy processing. To ensure diversity and accuracy, we designed two data production pipelines. The first one uses various VLMs, such as Geinimi, Qwen, and HunYuan, to generate detailed video captions from predefined prompts. Large Language Models (LLMs) like GPT-4, Gemini, and Qwen72B then use these captions and additional prompts to create question-answer pairs. The second approach directly utilizes VLMs to generate question-answer pairs from videos and prompts. The outputs from both pipelines are then combined into a comprehensive question-answer pool.

\begin{figure}[tp]
	\begin{minipage}[t]{1.0\linewidth}%\centering
	\includegraphics[width=0.9\linewidth]{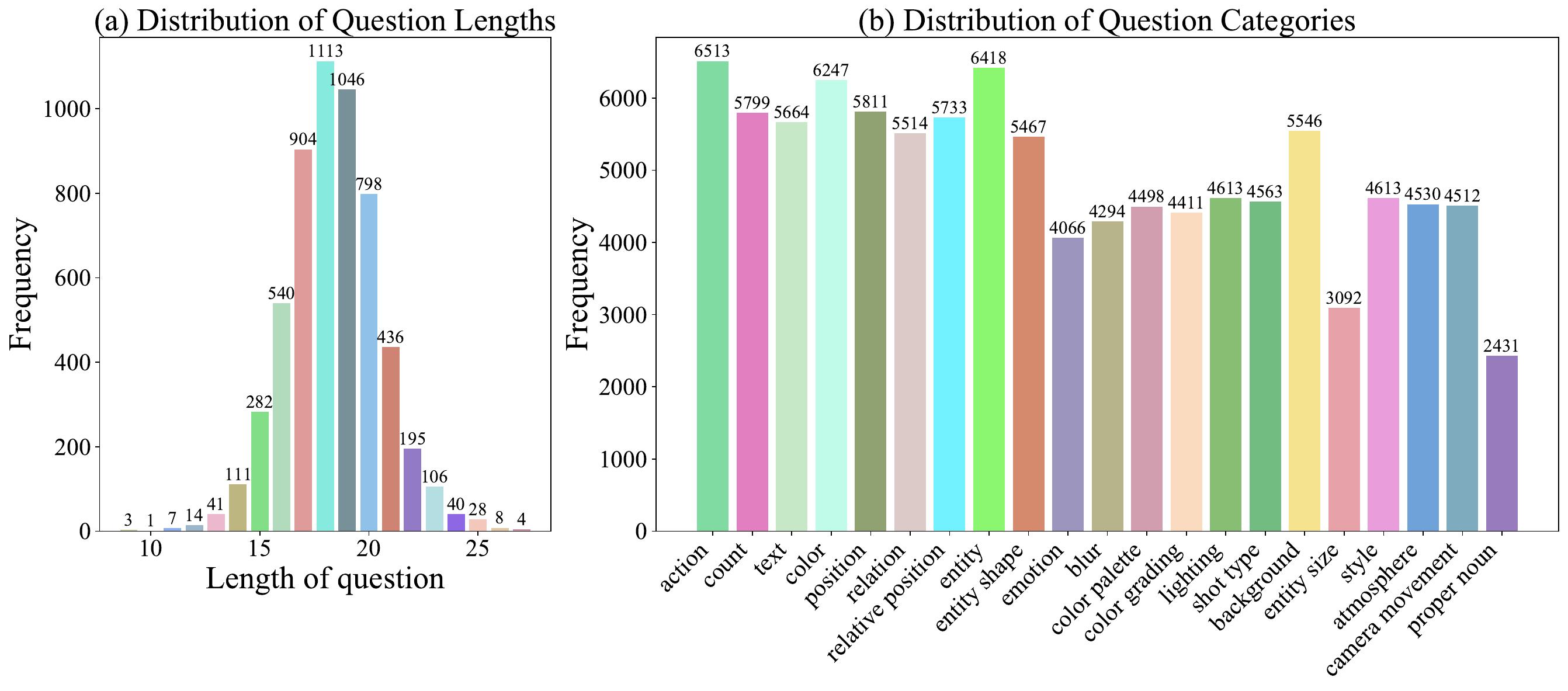}
	\caption{The question length and category distribution.}
	\label{fig:fig5}
	\end{minipage}%
	\vspace{-1.0em}	
\end{figure}

To enhance the quality of the QA-pairs, we established a data correction pipeline. This pipeline takes multiple sets of QA-pairs and captions from the same video as input and uses an advanced LLM (Gemini1.5) with predefined instructions to de-duplicate, filter, and retain high-quality QA-pairs. The instructions guide the LLM to merge similar QA-pairs, filter out those with the same question but different answers, and remove QA-pairs that appear only once, as they are likely to be invalid.

\textbf{Human Correction.} After generating high-quality candidate QA-pairs, we sample the data according to latitude to ensure that the data focuses more on important latitudes, and then conduct a final quality control process manually. Human reviewers re-examine these QA-pairs, deleting those with unreasonable or incorrect questions and correcting those with erroneous answers. Following this manual revision, we establish a benchmark (VCapsBench) consisting of 5,677 videos and 109,796 question-answer pairs. As shown in Fig.~\ref{fig:fig5}, each video contains between 10 and 27 QA-pairs, with each category comprising between 2,431 and 6,513 QA-pairs. The questions are organized into four major categories and 21 subcategories. Most answers are ``yes" providing a clear indication of correctness, while a smaller number of ``no" answers help assess whether the video captions contain hallucinations or inaccuracies.

\begin{figure*}[tp]
	\begin{minipage}[t]{1.0\linewidth}
        \centering
	\includegraphics[width=1.0\linewidth]{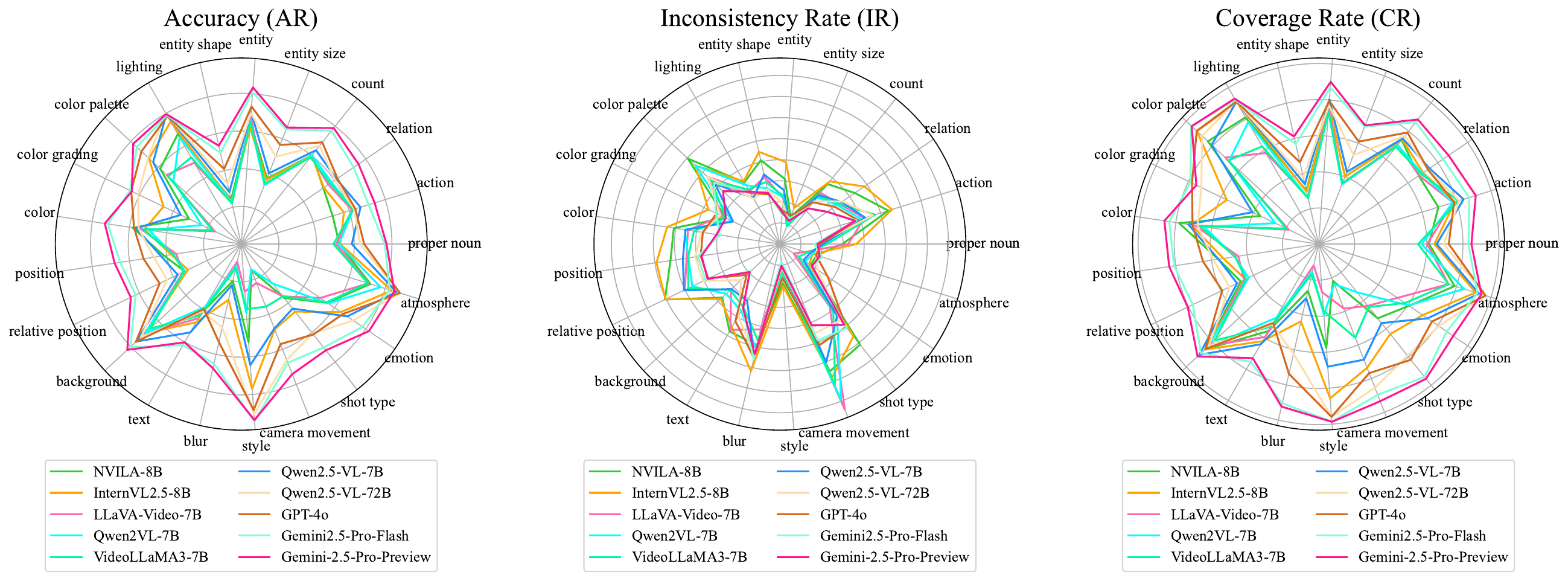}
	\caption{Results of Gemini-2.5-Pro-Preview captioning evaluation, organized by category.}
	\label{fig:fig6}
	\end{minipage}%
	\vspace{-1.5em}	
\end{figure*}

\subsection{Captioning Evaluation}
We employ VCapsBench to evaluate video captions produced by various vision-language models (VLMs), including top proprietary models like Gemini-1.5. For each test video, an evaluation-capable VLM generates a detailed caption according to specific guidelines and an output schema. This caption, along with each question from the video's QA pairs, is fed into a Large Language Model (LLM). As shown in Fig.~\ref{fig:fig1}, the LLM responds to each question based on the caption, providing answers in the format ``[answer], [reason]." We instruct the LLMs to assess the caption in the following three scenarios:

\begin{itemize}
    \item \textbf{Positive:} The caption accurately describes the relevant content, and the LLM's response aligns with the answer.
    
    \item \textbf{Negative:} The caption mentions the relevant content, but the LLM's response does not align with the answer.
    
    \item \textbf{Unanswerable:} The caption does not involve the relevant content for the dimension.
\end{itemize}

To comprehensively evaluate the quality of the captions, we developed three metrics tailored to these scenarios. The quality of the captions is evaluated using the following measure:

\begin{equation}
AR = \dfrac{N(\text{Positive})}{N(\text{All})},
\end{equation}

\begin{equation}
IR = \dfrac{N(\text{Negative})}{N(\text{Positive}) + N(\text{Negative})},
\end{equation}

\begin{equation}
CR = \dfrac{N(\text{Positive}) + N(\text{Negative})}{N(\text{All})},
\end{equation}

\textbf{Accuracy (AR):} This metric evaluates the percentage of ``Positive" responses, indicating that the caption correctly describes the relevant content and aligns with the LLM's response. A higher AR signifies a more accurate and reliable caption, reflecting better quality in the LLM's outputs.

\textbf{Inconsistency Rate (IR):} This metric measures the percentage of ``Negative" responses among all responses that reference the relevant content (both ``Positive" and ``Negative"). A lower IR denotes a more accurate caption, indicating that when the caption involves relevant content, it is more likely to be consistent with the real video content.

\textbf{Coverage Rate (CR):} This metric evaluates the total percentage of "Positive" and ``Negative" responses, reflecting whether the caption contains the relevant content, independent of the LLM's response consistency. A higher CR indicates a superior caption richness, as it suggests that the caption encompasses a greater amount of pertinent content from the video.

\begin{table*}[!t]
\renewcommand{\arraystretch}{0.6}
\caption{The accuracy (AR), inconsistency rate (IR), and coverage rate (CR) of VLM methods on all dimensions, where Gemini-2.5-Pro-Preview as a TextQA expert. The symbol ``$\uparrow$"  indicates that the larger the value, the better; The symbol ``$\downarrow$" indicates that the smaller the value, the better.}
\label{tab:results_1}
\resizebox{\textwidth}{!}{%
\begin{tabular}{@{}ll|cccccccccccc@{}}
\toprule
 &\multirow{3}{*}{Methods}& \multicolumn{7}{c|}{Content \& Entity}& \multicolumn{4}{c}{Color \&  Light}\\
 \cline{3-13} 
 &  & \begin{tabular}[c]{@{}c@{}}\small Proper\\ \small Noun\end{tabular} & \begin{tabular}[c]{@{}c@{}}Action\end{tabular} & \begin{tabular}[c]{@{}c@{}}Relation
\end{tabular} & \begin{tabular}[c]{@{}c@{}}Count
\end{tabular} & Entity  & \begin{tabular}[c]{@{}c@{}}\small Entity\\ \small Size\end{tabular} & \multicolumn{1}{c|}{\begin{tabular}[c]{@{}c@{}}\small Entity\\ \small Shape\end{tabular} }
&\begin{tabular}[c]{@{}c@{}}Lighting
\end{tabular} & \begin{tabular}[c]{@{}c@{}}\small Color \\ \small Palette \end{tabular} & \begin{tabular}[c]{@{}c@{}}\small Color\\ \small Grading
\end{tabular} & \begin{tabular}[c]{@{}c@{}}Color
\end{tabular}  \\ \midrule
\multirow{8}{*}{\rotatebox{90}{AR}$\Bigg\uparrow$} 
% & Qwen2VL-72B & 36.53 & 43.15 & 38.62 & 42.72 & 49.51 & 22.29 & \multicolumn{1}{c|}{14.45} & 50.81 & 37.25 & 20.20 &35.08  \\
& LLaVA-Video-7B & 50.45 & 61.6 & 55.98 & 59.47 & 67.68 & 34.22 & \multicolumn{1}{c|}{22.27} & 49.83 & 53.92 & 15.52 &\multicolumn{1}{c|}{53.96} \\
& Qwen2VL-7B & 51.98 & 62.11 & 58.08 & 60.27 & 65.35 & 33.60 & \multicolumn{1}{c|}{22.62} & 65.38 & 49.11 & 23.33 &\multicolumn{1}{c|}{49.49} \\ 
& VideoLLaMA3-7B & 48.97 & 66.29 & 57.12 & 58.94 & 65.30 & 34.45 & \multicolumn{1}{c|}{22.07} & 53.14 & 52.92 & 35.67 &\multicolumn{1}{c|}{56.48}\\
& NVILA-8B & 52.12 & 50.77 & 52.74 & 60.39 & 64.90 & 37.36 & \multicolumn{1}{c|}{24.21} & 67.37 & 58.71 & 30.74 &\multicolumn{1}{c|}{58.27} \\ 
& InternVL2.5-8B & 52.55 & 57.07 & 53.10 & 60.18 & 62.07 & 36.49 & \multicolumn{1}{c|}{23.04} & 74.97 & 66.81 & 45.82 &\multicolumn{1}{c|}{52.04} \\
& Qwen2.5VL-7B & 58.73 & 66.29 & 61.23 & 63.69 & \textbf{70.13} & 40.38 & \multicolumn{1}{c|}{28.28} & 79.58 & 63.81 & 35.67 &\multicolumn{1}{c|}{56.48} \\ 
& Qwen2.5-VL-72B & 63.14 &62.77 &60.25 &66.74 & 49.81 &69.71& \multicolumn{1}{c|}{32.72} & \textbf{79.99} &67.46 &60.77 &\multicolumn{1}{c|}{55.09} \\ 

& GPT-4o &65.28 &63.39 &61.81 &69.04 &56.51 &73.16 &\multicolumn{1}{c|}{40.9} &78.41 &72.27 &64.86 &\multicolumn{1}{c|}{57.56} \\ 
& Gemini2.5-Pro-Flash &76.19 &70.76 &72.23 &76.94 &65.26 &80.91 & \multicolumn{1}{c|}{49.66} &79.01 &75.68 &\textbf{65.56} &\multicolumn{1}{c|}{71.65} \\ 

& Gemini-2.5-Pro-Preview &\textbf{77.14} &\textbf{74.13} &\textbf{75.35} &\textbf{78.72} &66.43 &\textbf{83.33} & \multicolumn{1}{c|}{\textbf{53.53}} &79.67 &\textbf{78.24} &64.47 &\multicolumn{1}{c|}{\textbf{73.27}} \\ 

% & Gemini-1.5 & \textbf{86.28} & \textbf{90.29} & \textbf{89.51} & \textbf{80.75} & \textbf{94.99} & \textbf{69.05} & \multicolumn{1}{c|}{\textbf{59.05}} & \textbf{96.20} & \textbf{97.39} & \textbf{74.90}&\textbf{86.28}\\ 
\midrule
\multirow{8}{*}{\rotatebox{90}{IR}$\Bigg\downarrow$} 
% & Qwen2VL-72B & 12.78 & 20.81 & 16.37 & 14.88 & 9.37 & 4.97 & \multicolumn{1}{c|}{14.25} & 13.79 & 25.39 & 11.37 & 22.29 \\
& LLaVA-Video-7B & 16.21 & 20.98 & 16.96 & 15.54 & 10.82 & 5.80 & \multicolumn{1}{c|}{16.32} & 14.44 & 23.46 & 15.37 & \multicolumn{1}{c|}{22.08}\\
& Qwen2VL-7B & 9.22 & 20.15 & 16.42 & 14.11 & 11.16 & 5.47 & \multicolumn{1}{c|}{13.52} & 16.14 & 27.44 & 12.60 & \multicolumn{1}{c|}{22.84} \\
& VideoLLaMA3-7B & 11.20 & 23.64 & 16.94 & 14.49 & 11.42 & \textbf{4.49} & \multicolumn{1}{c|}{15.09} & 14.95 & 22.30  & 14.25 & \multicolumn{1}{c|}{20.62} \\
& NVILA-8B & 14.81 & 26.92 & 21.50 & 18.19 & 15.49 & 6.94  & \multicolumn{1}{c|}{20.37} &16.89 & 29.80 & 14.37 & \multicolumn{1}{c|}{25.46} \\
& InternVL2.5-8B & 18.10 & 27.82 & 24.24 & 19.00 & 19.50 & 9.19 & \multicolumn{1}{c|}{22.42} & 17.18 & 27.38 & 18.80 & \multicolumn{1}{c|}{27.06} \\
& Qwen2.5VL-7B & 10.26 & 21.20 & 17.72 & 14.86 & 12.81 & 6.03 & \multicolumn{1}{c|}{16.95} & 12.60 &  21.00 & \textbf{12.34} & \multicolumn{1}{c|}{22.65} \\

& Qwen2.5-VL-72B &8.97 &23.08 &17.97 &12.94 &5.99 &9.58 & \multicolumn{1}{c|}{13.35} &\textbf{12.35} &23.47 &15.69 &\multicolumn{1}{c|}{18.76} \\ 
& GPT-4o &9.53 &19.36 &15.71 &12.66 &7.28 &7.96 & \multicolumn{1}{c|}{\textbf{12.1}} &13.95 &21.54 &16.69 &\multicolumn{1}{c|}{18.5} \\ 
& Gemini2.5-Pro-Flash &\textbf{8.23} &19.04 &\textbf{13.55} &10.91 &6.63 &7.0 & \multicolumn{1}{c|}{12.56} &14.22 &21.4 &16.83 &\multicolumn{1}{c|}{\textbf{14.4}}  \\

& Gemini-2.5-Pro-Preview &8.95 &\textbf{18.7} &13.59 &\textbf{10.79} &\textbf{5.92} &7.54 & \multicolumn{1}{c|}{12.42} &14.36 &\textbf{18.42} &14.23 &\multicolumn{1}{c|}{15.21}\\

% & Gemini-1.5 & \textbf{4.73} & \textbf{8.18} & \textbf{7.02} & \textbf{9.02} & \textbf{3.89} &\textbf{3.75}  & \multicolumn{1}{c|}{\textbf{7.07}} & \textbf{3.17} & \textbf{2.32} & \textbf{6.71} & \textbf{8.71}\\ 
 \midrule
\multirow{8}{*}{\rotatebox{90}{CR}$\Bigg\uparrow$} 
% & Qwen2VL-72B & 41.89 & 54.48 & 46.18 & 50.19 & 54.63 & 23.46 & \multicolumn{1}{c|}{16.85} & 58.94 & 49.92 & 22.79 & 45.14  \\
& LLaVA-Video-7B & 60.21 & 77.95 & 67.41 & 70.41 & 75.89 & 36.33 & \multicolumn{1}{c|}{26.62} & 58.24 & 70.45 & 18.33 &\multicolumn{1}{c|}{69.26}  \\
& Qwen2VL-7B & 57.17 & 77.79 & 69.50 & 70.20 & 73.56 & 35.55 & \multicolumn{1}{c|}{26.26} & 77.97 & 67.68 & 26.69 &\multicolumn{1}{c|}{64.15} \\ 
& VideoLLaMA3-7B & 55.14 & 81.34 & 68.77 & 68.92 & 73.72 & 36.07 & \multicolumn{1}{c|}{26.00} & 62.48 & 68.10 & 19.76 &\multicolumn{1}{c|}{66.53}  \\
& NVILA-8B & 61.19 & 69.48 & 67.19 & 73.82 & 76.80 & 40.15 & \multicolumn{1}{c|}{30.41} & 81.06 & 83.63 & 35.89 &\multicolumn{1}{c|}{78.17} \\
& InternVL2.5-8B & 64.17 & 79.06 & 70.08 & 74.30 & 77.11 & 40.15 & \multicolumn{1}{c|}{29.68} & 90.52 & 92.00 & 56.43 &\multicolumn{1}{c|}{71.34}  \\
& Qwen2.5VL-7B & 65.44 & 84.12 & 74.43 & 74.80 & \textbf{80.44} & 42.97 & \multicolumn{1}{c|}{34.06} & 91.05 & 80.78 & 40.69 & \multicolumn{1}{c|}{73.03}\\ 

& Qwen2.5-VL-72B &69.36 &81.6 &73.45 &76.66 &52.98 &77.1 & \multicolumn{1}{c|}{37.76} &91.27 &88.15 &72.08 &\multicolumn{1}{c|}{67.81} \\ 

& GPT-4o &72.16 &78.61 &73.34 &79.05 &60.95 &79.48 & \multicolumn{1}{c|}{46.53} &91.11 &92.12 &77.85 &\multicolumn{1}{c|}{70.62} \\ 
& Gemini2.5-Pro-Flash &83.03 &87.4 &83.55 &86.36 &69.9 &86.99 &\multicolumn{1}{c|}{56.79} &92.11 &\textbf{96.28} &\textbf{78.83} &\multicolumn{1}{c|}{83.7} \\ 
& Gemini-2.5-Pro-Preview &\textbf{84.72} &\textbf{91.17} &\textbf{87.2} &\textbf{88.25} &70.61 &\textbf{90.12} &\multicolumn{1}{c|}{\textbf{61.12}} &\textbf{93.03} &95.9 &75.17 &\multicolumn{1}{c|}{\textbf{86.41}} \\

% & Gemini-1.5 & \textbf{90.56} & \textbf{98.33} & \textbf{96.27} & \textbf{88.75} & \textbf{98.82} & \textbf{71.74} & \multicolumn{1}{c|}{\textbf{63.54}} & \textbf{99.34} & \textbf{99.71} & \textbf{80.28} &\textbf{94.51} \\ 
 \bottomrule
 \toprule
 &\multirow{3}{*}{Methods}& \multicolumn{5}{c|}{Visuals \& Composition}& \multicolumn{5}{c|}{Cinematography  \& Environment}\\
 \cline{3-13} 
 &  & Position & \begin{tabular}[c]{@{}c@{}}\small Relative\\ \small Position\end{tabular} & Background &Text & \multicolumn{1}{c|}{Blur} & Style & \begin{tabular}[c]{@{}c@{}}\small Camera \\ \small Movement\end{tabular} & \begin{tabular}[c]{@{}c@{}}Shot Type\end{tabular} & \begin{tabular}[c]{@{}c@{}}Emotion\end{tabular} & \multicolumn{1}{c|}{Atmosphere} & \color{red}\textbf{ALL}\\ \midrule
\multirow{8}{*}{\rotatebox{90}{AR}$\Bigg\uparrow$} 
% & Qwen2VL-72B & 22.85 & 21.29 & 48.67 & 26.77 & \multicolumn{1}{c|}{10.43} & 35.10 & 16.17 & 21.71 & 35.94 & \multicolumn{1}{c|}{52.14} &33.01 \\ 
& LLaVA-Video-7B & 34.91 & 34.31 & 69.57 & 45.47 & \multicolumn{1}{c|}{9.61} & 25.30 & 22.17 & 35.47 & 50.78 & \multicolumn{1}{c|}{70.64} &45.11 \\
& Qwen2VL-7B & 36.69 & 33.10 & 69.45 & 38.51 & \multicolumn{1}{c|}{11.72} & 36.86 & 14.33 & 26.76 & 55.96 & \multicolumn{1}{c|}{77.41} &45.75\\ 
& VideoLLaMA3-7B & 36.45 & 33.80 & 65.86 & 43.73 & \multicolumn{1}{c|}{13.09} & 34.51 & 35.75 & 35.51 & 54.84 & \multicolumn{1}{c|}{70.33} &46.13 \\
& NVILA-8B & 41.44 & 33.31 & 69.79 & 42.32 & \multicolumn{1}{c|}{20.04} & 52.29 & 14.81 & 36.54 & 55.36 & \multicolumn{1}{c|}{71.74} &47.99  \\
& InternVL2.5-8B & 37.55 & 31.37 & 69.44 & 47.99 & \multicolumn{1}{c|}{30.36} & 77.08 & 48.61 & 45.90 & 64.00 & \multicolumn{1}{c|}{83.67} &53.30  \\
 & Qwen2.5VL-7B & 43.06 & 37.43 & 76.03 & 54.13 & \multicolumn{1}{c|}{22.33} & 64.43 & 48.10 & 43.79 & 68.12 & \multicolumn{1}{c|}{86.17} &56.03 \\ 
& Qwen2.5-VL-72B & 46.93 &43.73 &72.59 &41.27 & \multicolumn{1}{c|}{45.37} & 90.53 &66.07 &60.76 &71.97 & \multicolumn{1}{c|}{87.94} &61.20 \\
& GPT-4o &52.45 &48.08 &76.35 &39.77 & \multicolumn{1}{c|}{53.33} &88.29 &57.03 &61.35 &65.64 & \multicolumn{1}{c|}{\textbf{88.16}} &63.08 \\

& Gemini2.5-Pro-Flash &65.03 &62.2 &80.47 &\textbf{61.95} &
\multicolumn{1}{c|}{65.09} &\textbf{94.48} &67.67 &70.17 &77.94 & \multicolumn{1}{c|}{82.5} &71.77 \\
& Gemini-2.5-Pro-Preview &\textbf{67.98} &\textbf{65.12} &\textbf{82.55} &60.19 & \multicolumn{1}{c|}{\textbf{67.56}} &93.63 &\textbf{74.12} &\textbf{72.08} &\textbf{82.03} & \multicolumn{1}{c|}{85.09} &\color{red}\textbf{73.88} \\

 % & Gemini-1.5 & \textbf{83.91} & \textbf{78.63} & \textbf{94.78} & \textbf{79.22} & \multicolumn{1}{c|}{\textbf{80.19}} &\textbf{98.02} & \textbf{77.41} & \textbf{81.89} & \textbf{95.76} & \multicolumn{1}{c|}{\textbf{85.93}} &\textbf{84.99} \\ 
 \midrule
\multirow{8}{*}{\rotatebox{90}{IR}$\Bigg\downarrow$} 
% & Qwen2VL-72B & 23.32 & 22.62 & 16.36 & 29.48 & \multicolumn{1}{c|}{27.79} & 7.13 & 33.58 & 21.57 & 7.25  & \multicolumn{1}{c|}{7.72} &17.04 \\
& LLaVA-Video-7B & 22.41 & 25.02 & 15.83 & 23.50 & \multicolumn{1}{c|}{\textbf{20.77}} & 5.52 & 42.21 & \textbf{18.96} & \textbf{3.96}  & \multicolumn{1}{c|}{\textbf{6.28}} & 17.46 \\
& Qwen2VL-7B & 21.94 & 23.07 & 17.36 & 18.34 & \multicolumn{1}{c|}{25.41} & 6.45 & 39.83 & 21.71 & 7.24 & \multicolumn{1}{c|}{8.21} &16.79 \\
& VideoLLaMA3-7B & 22.05 & 24.13 & 16.06 & 17.30 & \multicolumn{1}{c|}{23.47} & 6.75 & 35.85 & 20.33 & 5.36  & \multicolumn{1}{c|}{6.83} & 17.09\\
& NVILA-8B & 26.26 & 30.28 & 18.66 & 23.91 & \multicolumn{1}{c|}{24.58} & 9.34 & 32.42 & 30.51 & 8.77 & \multicolumn{1}{c|}{9.11} &20.45\\
& InternVL2.5-8B & 29.63 & 30.16 & 19.15 & 22.74 & \multicolumn{1}{c|}{30.91} & 10.04 & 33.98 & 28.00 & 8.40 & \multicolumn{1}{c|}{8.14} &21.63  \\
& Qwen2.5VL-7B & 23.29 & 25.29 & 15.68 & \textbf{15.54} & \multicolumn{1}{c|}{27.39} & 5.39 & 30.07 & 21.80 &  6.78& \multicolumn{1}{c|}{7.46} & 16.65 \\

& Qwen2.5-VL-72B &20.32 &20.24 &12.44 &19.71 & \multicolumn{1}{c|}{27.16} &5.6 &22.71 &23.11 &10.76  & \multicolumn{1}{c|}{9.06} &16.02\\
& GPT-4o &19.01 &\textbf{18.87} &10.73 &21.12 & \multicolumn{1}{c|}{27.69} &8.01 &25.72 &25.19 &13.74 &\multicolumn{1}{c|}{9.02} &15.90 \\

&  Gemini2.5-Pro-Flash &19.71 &19.61 &\textbf{9.49} &17.69 &
\multicolumn{1}{c|}{28.04} &\textbf{4.14} &24.53 &25.48 &9.29 &\multicolumn{1}{c|}{10.47}& 15.14 \\

&Gemini-2.5-Pro-Preview &\textbf{18.8} &19.05 &9.76 &17.48 &\multicolumn{1}{c|}{26.82} &5.15 &\textbf{20.72} &24.51 &9.2 & \multicolumn{1}{c|}{9.89} & \color{red}\textbf{14.62}\\

% & Gemini-1.5 & \textbf{10.34} & \textbf{13.33} & \textbf{3.78} & \textbf{10.38} & \multicolumn{1}{c|}{\textbf{9.79}} & \textbf{1.18} & \textbf{20.78} & \textbf{15.50} & \textbf{1.70} & \multicolumn{1}{c|}{\textbf{6.93}} &\textbf{7.71} \\ 
 \midrule
\multirow{8}{*}{\rotatebox{90}{CR}$\Bigg\uparrow$} 
% & Qwen2VL-72B & 29.80 & 27.51 & 58.20 & 37.97 & \multicolumn{1}{c|}{14.45} & 37.80 & 24.35 & 27.68 & 38.75 & \multicolumn{1}{c|}{56.49} &39.79  \\ 
& LLaVA-Video-7B & 45.00 & 45.75 & 82.27 & 59.44 & \multicolumn{1}{c|}{12.14} & 26.78 & 73.63 & 63.74 & 52.87 & \multicolumn{1}{c|}{75.38} &54.65 \\
& Qwen2VL-7B & 47.00 & 43.03 & 84.04 & 47.16 & \multicolumn{1}{c|}{15.71} & 39.40 & 23.82 & 34.17 & 60.32 & \multicolumn{1}{c|}{84.33} &54.98 \\ 
& VideoLLaMA3-7B & 46.77 & 44.55 & 78.46 & 64.08 & \multicolumn{1}{c|}{17.11} & 37.01 & 55.71 & 55.99 & 57.95 & \multicolumn{1}{c|}{75.49} & 55.64  \\
& NVILA-8B & 56.21 & 47.78 & 85.80 & 55.62 & \multicolumn{1}{c|}{26.58} & 57.68 & 21.93 & 52.59 & 60.69 & \multicolumn{1}{c|}{78.94} &60.33  \\
& InternVL2.5-8B & 53.36 & 44.91 & 85.89 & 62.12 & \multicolumn{1}{c|}{43.94} & 85.68 & 21.93 & 52.59 & 60.68 & \multicolumn{1}{c|}{91.08} &68.00 \\
& Qwen2.5VL-7B & 56.13 & 50.10 & 90.17 & 64.08 & \multicolumn{1}{c|}{14.45} & 37.80 & 24.35 & 27.68 & 38.74 & \multicolumn{1}{c|}{93.12} &67.22  \\

& Qwen2.5-VL-72B &58.9 &54.82 &82.91 &51.41 &\multicolumn{1}{c|}{62.29} &95.89 &85.49 &79.02 &80.65 & \multicolumn{1}{c|}{96.7} &72.87\\
& GPT-4o  &64.76 &59.26 &85.53 &50.42 &\multicolumn{1}{c|}{73.75} &95.98 &76.78 &82.0 &76.09 &\multicolumn{1}{c|}{\textbf{96.9}} &75.01 \\
& Gemini2.5-Pro-Flash &81.0 &77.37 &88.91 &\textbf{75.26} &
\multicolumn{1}{c|}{90.46} &98.57 &89.67 &94.16 &85.93 &\multicolumn{1}{c|}{ 92.15} & 84.57\\
& Gemini-2.5-Pro-Preview &\textbf{83.72} &\textbf{80.44}&\textbf{91.47} &72.94 &\multicolumn{1}{c|}{\textbf{92.32}}&\textbf{98.72} &\textbf{93.49} &\textbf{95.48} &\textbf{90.34} & \multicolumn{1}{c|}{94.42} &\color{red}\textbf{86.52} \\

% & Gemini-1.5 &  \textbf{93.59} & \textbf{90.72} & \textbf{98.50} & \textbf{88.40} & \multicolumn{1}{c|}{\textbf{88.90}} & \textbf{99.20}  & \textbf{97.71} & \textbf{96.90} & \textbf{97.41} & \multicolumn{1}{c|}{\textbf{92.32}} &\textbf{92.09} \\ 

\bottomrule
\end{tabular}%
}
\vspace{-1.0em}
\end{table*}

\section{Experiment}
\textbf{Experimental Setup.} We evaluated several popular open-source vision-language models (VLMs), including Qwen2VL~\cite{wang2024qwen2}, Qwen2.5VL~\cite{bai2025qwen2}, InternVL2.5~\cite{bai2025qwen2}, LLaVA-Video~\cite{zhang2024videoinstructiontuningsynthetic}, NVILA~\cite{liu2024nvila}, and VideoLLaMA3~\cite{zhang2025videollama}, alongside the closed-source Gemini-2.5 and GPT-4o accessed via API. All models were prompted identically to generate detailed video descriptions across 21 dimensions. For better evaluation, we adopted two  powerful LLM (\eg, Gemini-2.5-Pro-Preview and GPT-4.1) as TextQA experts to answer VCapsBench questions based on the generated descriptions. The evaluation metrics, derived from their responses, are presented in Table.~\ref{tab:results_1} and Appendix~\ref{exp_app} Table.~\ref{tab:results_2}. Following the Playground v3 protocol~\cite{playgroundv3}, each caption was queried three times, and a consensus response was obtained to minimize output variability and ensure consistency.

\begin{figure}[thbp]
	\begin{minipage}[t]{1.0\linewidth}
        \centering
	\includegraphics[width=1.0\linewidth]{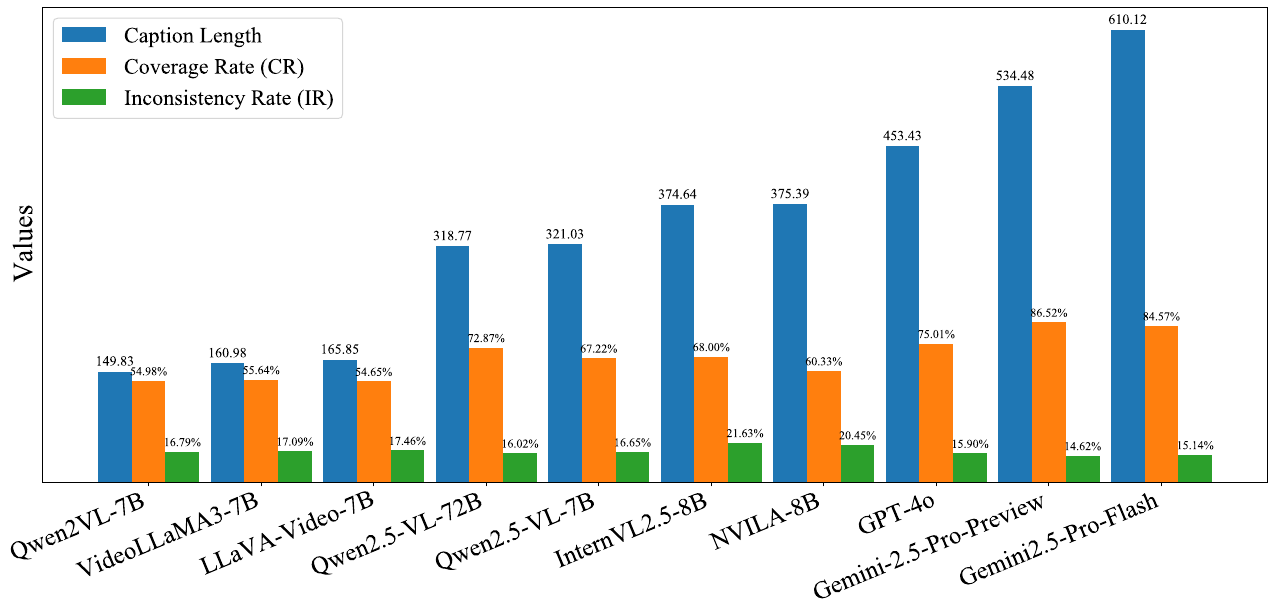}
    \vspace{-0.5em}
	\caption{The relationship between CR, IR and caption length, where Gemini-2.5-Pro-Preview as a TextQA expert. Caption length is the number of words in the caption, not counting special symbols.}
	\label{fig:fig7}
	\end{minipage}%
	\vspace{-1.5em}	
\end{figure}

\subsection{Main Results}

\textbf{Accuracy (AR):} As illustrated in Table~\ref{tab:results_1} and Fig.~\ref{fig:fig6}, Gemini-2.5-Pro-Preview consistently surpasses all other models across various dimensions when evaluated using the Accuracy (AR) metric. For instance, in the ``Content \& Entity" dimension, specifically concerning action, Gemini-2.5-Pro-Preview achieves an AR of 74.13\%, whereas other models typically fall between 40\% and 70\%. In the ``Color \& Light" dimension, particularly for lighting, Gemini-2.5-Pro-Preview records an impressive AR of 79.67\%, significantly outperforming its counterparts. When considering the entire dataset, Gemini-2.5-Pro-Preview achieves an AR of 73.88\%, which is markedly superior to that of other open-source models. Among these, the latest Qwen2.5-VL-72B shows the best performance with an AR of 61.20\%, yet it remains considerably lower than Gemini-2.5-Pro-Preview. This underscores Gemini-2.5-Pro-Preview's exceptional ability to accurately capture and describe video content, thereby delivering higher quality descriptions and insights.

\textbf{Inconsistency Rate (IR):} Gemini-2.5-Pro-Preview stands out in terms of the IR metric, achieving the lowest IR values. For instance, in the ``Visuals \& Composition'' category, especially for background, Gemini-2.5-Pro-Preview records an IR of just 9.76\%, while other models typically range between 10\% and 20\%. In the ``Cinematography \& Environment'' category, especially for lighting, Gemini-2.5-Pro-Flash's IR is only 14.22\%.

\textbf{Coverage Rate (CR):} When it comes to the CR metric, Gemini-2.5-Pro-Preview again shows superior performance. In the ``Content \& Entity'' category, the lowest CR for Gemini-2.5-Pro-Preview is 61.12\% in the ``Entity Shape'' subcategory, while the highest is 90.12\% in the ``Entity'' subcategory. Other open-source models have CR values that are more than 10\% lower across all categories. Overall, Gemini-2.5-Pro-Preview achieves an impressive CR of 86.52\%. Among open-source models, InternVL2.5-8B performs best with a CR of 68.00\%, which is 18.52\% lower than Gemini-2.5-Pro-Preview. This demonstrates that Gemini-2.5-Pro-Preview can cover more relevant content from the video, providing more comprehensive coverage regardless of description consistency. 

Gemini-2.5-Pro-Preview excels in generating video descriptions, demonstrating high accuracy (AR), consistency (IR), and comprehensive coverage (CR). This underscores its superior performance in producing high-quality, detailed, and consistent descriptions compared to other models. VCapsBench highlights the shortcomings of existing open-source VLMs, such as the lack of detailed object shapes, sizes, colors, and lighting in descriptions. These deficiencies can be particularly problematic for applications requiring highly detailed and complete captions, such as text-to-video generation models and advanced video analysis tools.

\subsection{Evaluation Analysis}
Although Gemini surpasses open-source VLMs in every aspect, showcasing its advanced video comprehension abilities, it is crucial to acknowledge that Gemini's captions were utilized in creating QA-pairs, potentially influencing its elevated CR. Nevertheless, Gemini's metrics are valuable as they establish a benchmark for open-source models, emphasizing the disparity in video content understanding. This insight is instrumental in guiding the optimization of open-source VLMs.

\begin{figure*}[thbp]
	\begin{minipage}[t]{1.0\linewidth}%\centering
	\includegraphics[width=1\linewidth]{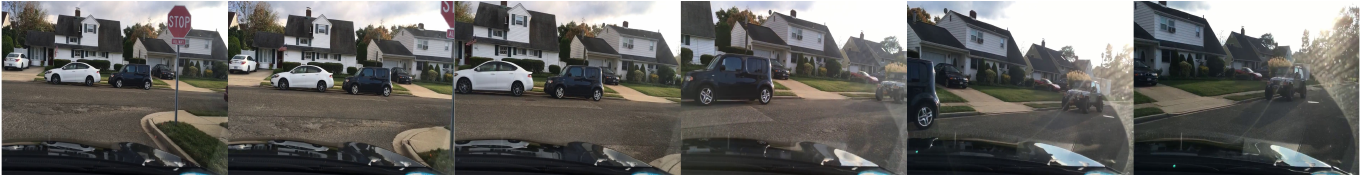}
    \vspace{-1.5em}
     \caption{Representative video from VCapsBench.}
	\label{fig:fig9}
	\end{minipage}%
	\vspace{-1.5em}	
\end{figure*}

\begin{table*}[htbp]
\centering
\setlength{\tabcolsep}{6pt}
\renewcommand{\arraystretch}{0.8}
\caption{Questions for representative video from VCapsBench. ``Q2.5-72B", ``LV-7B", ``Q2-7B", ``VL-7B", ``NV-8B", ``I2.5-8B", ``Q2.5-7B", ``G2.5" represent ``Qwen2.5VL-72B", ``LLaVA-Video-7B", ``Qwen2VL-7B", ``VideoLLaMA3-7B", ``NVILA-8B", ``InternVL2.5-8B", ``Qwen2.5VL-7B" and ``Gemini-2.5-Pro-Preview" respectively. ``\CheckmarkBold", ``\XSolidBrush", ``\boldmath $\bullet$" represent ``Positive", ``Negative" and ``Unanswerable", respectively.}
\resizebox{\textwidth}{!}{%
\begin{tabular}{l|l|p{9.2cm}|c|cccccccc}
\toprule 
    % &Category & Question & GT & \begin{tabular}[c]{@{}c@{}}\small Qwen2VL\\ \small 72B\end{tabular} & 
    % \begin{tabular}[c]{@{}c@{}}\small LLaVA\\ \small Video-7B\end{tabular} & \begin{tabular}[c]{@{}c@{}}\small Qwen2VL\\ \small 7B\end{tabular} & \begin{tabular}[c]{@{}c@{}}\small VideoLLaMA3\\ \small 7B\end{tabular}
    % & \begin{tabular}[c]{@{}c@{}}\small NVILA\\ \small 8B\end{tabular} & \begin{tabular}[c]{@{}c@{}}\small InternVL2.5\\ \small 8B\end{tabular}& \begin{tabular}[c]{@{}c@{}}\small Qwen2.5VL\\ \small 7B\end{tabular} & \small Gemini-1.5\\ 
    &Category & Question & GT & \small Q2.5-72B &\small LV-7B& \small Q2-7B & \small VL-7B
    & \small NV-8B & \small I2.5-8B & \small Q2.5-7B & \small G2.5\\ 
    \midrule
    \multirow{7}{*}{\rotatebox{90}{\begin{tabular}[c]{@{}c@{}}\small Content\\ \small  \& Entity \end{tabular} }}
    &action & Is there a subtle panning movement from left to right in the video? & yes & \CheckmarkBold & \XSolidBrush & \XSolidBrush & \CheckmarkBold & \boldmath $\bullet$ & \CheckmarkBold & \CheckmarkBold & \CheckmarkBold\\
    \cline{2-4} 
    &count & Are there more than three cars visible in the video? & yes & \boldmath $\bullet$ &  \CheckmarkBold &  \CheckmarkBold & \CheckmarkBold & \CheckmarkBold & \CheckmarkBold & \CheckmarkBold & \CheckmarkBold\\
    \cline{2-4} 
    &relation & Is the red car parked on the opposite side of the street? & yes & \boldmath $\bullet$ &  \boldmath $\bullet$& \CheckmarkBold  & \boldmath $\bullet$ & \boldmath $\bullet$ &\XSolidBrush &\boldmath $\bullet$ &\XSolidBrush\\
    \cline{2-4} 
    &entity& Is there a stop sign in the video? & yes &  \CheckmarkBold & \CheckmarkBold & \CheckmarkBold & \CheckmarkBold &\CheckmarkBold &\CheckmarkBold &\CheckmarkBold &\CheckmarkBold\\
    \cline{2-4} 
    &entity size & Is the black SUV larger than the white sedan? & yes & \boldmath $\bullet$ &  \boldmath $\bullet$ & \boldmath $\bullet$ &\boldmath $\bullet$  &\boldmath $\bullet$ &\boldmath $\bullet$ &\boldmath $\bullet$ &\boldmath $\bullet$\\
    \cline{2-4} 
    &entity shape& Is the stop sign octagonal? & yes & \boldmath $\bullet$ & \boldmath $\bullet$ & \CheckmarkBold & \boldmath $\bullet$ &\boldmath $\bullet$ & \boldmath $\bullet$ &\boldmath $\bullet$ & \CheckmarkBold\\
     \midrule
    \multirow{6}{*}{\rotatebox{90}{\begin{tabular}[c]{@{}c@{}}\small Color\\ \small  \& Light \end{tabular}}}
    &lighting & \begin{tabular}[l]{@{}l@{}}Is the lighting in video primarily natural and even, \\ characteristic of daytime? \end{tabular}& yes &  \CheckmarkBold & \boldmath $\bullet$ & \XSolidBrush & \boldmath $\bullet$ & \CheckmarkBold & \CheckmarkBold & \CheckmarkBold &  \CheckmarkBold \\
    \cline{2-4} 
    &color palette & Does the video have a realistic color palette? & yes & \CheckmarkBold &  \CheckmarkBold &\boldmath $\bullet$ & \boldmath $\bullet$ & \boldmath $\bullet$ &\CheckmarkBold &\CheckmarkBold &  \CheckmarkBold\\
    \cline{2-4} 
    &color grading & Is the image ungraded, with natural tones for the houses and cars? & yes & \boldmath $\bullet$ &  \CheckmarkBold &\boldmath $\bullet$ & \boldmath $\bullet$ & \boldmath $\bullet$ &\CheckmarkBold & \boldmath $\bullet$  &  \CheckmarkBold \\
    \cline{2-4} 
    &color &Is the white sedan parked directly in front of the stop sign? & yes & \boldmath $\bullet$ &  \boldmath $\bullet$ &  \CheckmarkBold &  \boldmath $\bullet$ & \XSolidBrush & \XSolidBrush & \boldmath $\bullet$ & \boldmath $\bullet$\\
    \midrule
    \multirow{6}{*}{\rotatebox{90}{\begin{tabular}[c]{@{}c@{}}\small Visuals \& \\ \small Composition \end{tabular}}}&position & Is the stop sign on the right side of the road? & yes & \boldmath $\bullet$ & \CheckmarkBold & \CheckmarkBold & \boldmath $\bullet$ & \CheckmarkBold & \CheckmarkBold &\boldmath $\bullet$ & \boldmath $\bullet$\\
    \cline{2-4} 
    &relative position & Is the white car closer to the stop sign than the black car? & no & \boldmath $\bullet$ &  \boldmath $\bullet$ & \XSolidBrush & \boldmath $\bullet$ &\boldmath $\bullet$ &\boldmath $\bullet$ & \boldmath $\bullet$ &\boldmath $\bullet$\\
    \cline{2-4} 
    &background & Is the video set in a residential suburban street? & yes & \CheckmarkBold & \CheckmarkBold & \CheckmarkBold & \CheckmarkBold & \CheckmarkBold & \CheckmarkBold & \CheckmarkBold &  \CheckmarkBold \\
    \cline{2-4} 
    &text & Is there a stop sign with the text ``ALL WAY" in the video? & yes & \CheckmarkBold & \XSolidBrush & \CheckmarkBold   & \CheckmarkBold & \CheckmarkBold & \XSolidBrush & \CheckmarkBold & \boldmath $\bullet$\\
    \cline{2-4} 
    &blur & Is there a shallow depth of field in the video? & yes & \boldmath $\bullet$ &  \CheckmarkBold &\boldmath $\bullet$ & \boldmath $\bullet$ & \boldmath $\bullet$ & \boldmath $\bullet$ & \boldmath $\bullet$ &  \CheckmarkBold \\
    \midrule
    \multirow{3}{*}{\rotatebox{90}{ \begin{tabular}[c]{@{}c@{}}\small Cinematography\\ \small  \& Environment \end{tabular}}}&camera movement & Is the camera movement in the video a slow panning from left to right? & yes & \CheckmarkBold & \XSolidBrush & \XSolidBrush & \CheckmarkBold & \boldmath $\bullet$ & \boldmath $\bullet$ & \XSolidBrush &  \CheckmarkBold\\
    \cline{2-4} 
    &shot type& Is the primary shot type predominantly a medium shot taken from a fixed perspective? & yes & \boldmath $\bullet$ & \XSolidBrush & \XSolidBrush & \boldmath $\bullet$ & \boldmath $\bullet$ & \XSolidBrush & \XSolidBrush &  \CheckmarkBold \\
    \cline{2-4} 
    &atmosphere & Is the video set in a calm and quiet atmosphere, typical of a residential neighborhood?  & yes & \CheckmarkBold & \CheckmarkBold  & \CheckmarkBold & \CheckmarkBold & \XSolidBrush & \XSolidBrush & \CheckmarkBold & \CheckmarkBold \\
\bottomrule
\end{tabular}
}
\label{tbl:bench_questions}
\vspace{-1.0em}	
\end{table*}

\textbf{Caption length distribution analysis.} We also explore the caption lengths to determine if longer captions are associated with a higher evaluation coverage rate (CR), as shown in Fig.~\ref{fig:fig7} . The Fig.~\ref{fig:fig11} in Appendix \ref{exp_app} presents histograms of caption word lengths for the models assessed. Our findings indicate that as the length of captions generated by models increases, the CR generally improves. However, this increase in length also leads to more errors in some open-source models, such as Qwen2VL-7B, VideoLLaMA3-7B, LLaVA-Video-7B, and InternVL2.5-8B. Interestingly, open-source VLMs like Qwen2.5-VL-72B and the closed VLM Gemini-2.5, despite producing longer captions, exhibit lower error rates compared to other VLMs. Additionally, Qwen2.5-VL-7B achieves a higher CR than VILA-8B with shorter captions and a lower IR than LLaVA-Video-7B, which also has shorter captions. This suggests that a deep understanding of video content allows models to generate concise, yet thorough and accurate descriptions.

\textbf{Representative sample analysis.} To further examine the strengths and weaknesses of various models, we conducted a qualitative analysis of captions. In Fig.~\ref{fig:fig9}, we present a video from VCapsBench, which features numerous subjects and intricate details, posing a significant challenge for VLMs. Table.~\ref{tbl:bench_questions} lists the questions for this video along with the answers derived from the captions. It is evident that different VLMs exhibit varying performances across different dimensions. Some categories, such as ``atmosphere'' and ``background'' are relatively easy for all VLMs, whereas others, like ``relative position'' present difficulties for all models. However, each video has unique questions across different categories. The comprehensive dataset includes a vast number of questions, sufficient to thoroughly evaluate a VLM's capability to understand and observe video details. More results in Appendix \ref{exp_app}, \ref{exp_app1} and \ref{exp_app2}.

\section{Conclusion}
In this work, we introduce VCapsBench, a new large-scale fine-grained benchmark for evaluating video caption quality. VCapsBench is meticulously designed to support detailed long captions, aiming to advance research and benchmarking in video understanding. The benchmark comprises over 5K videos and more than 100K QA-pairs, assessing 21 critical dimensions of video generation. We have evaluated the caption quality produced by various open-source and closed-source models using this benchmark. The comprehensive analysis highlights the strengths and weaknesses of these models in generating accurate and detailed captions. We believe that VCapsBench will play a crucial role in guiding the optimization of video caption generation, thereby advancing the development of text-to-video models and enhancing the overall understanding of video content.

\clearpage
\bibliography{
neurips_2025
}

\begin{thebibliography}{10}

\bibitem{reid2024gemini}
Machel Reid, Nikolay Savinov, Denis Teplyashin, Dmitry Lepikhin, Timothy Lillicrap, Jean-baptiste Alayrac, Radu Soricut, Angeliki Lazaridou, Orhan Firat, Julian Schrittwieser, et~al.
\newblock Gemini 1.5: Unlocking multimodal understanding across millions of tokens of context.
\newblock {\em arXiv preprint arXiv:2403.05530}, 2024.

\bibitem{liu2025st}
Ruyang Liu, Chen Li, Haoran Tang, Yixiao Ge, Ying Shan, and Ge~Li.
\newblock St-llm: Large language models are effective temporal learners.
\newblock In {\em European Conference on Computer Vision}, pages 1--18. Springer, 2025.

\bibitem{hong2024cogvlm2}
Wenyi Hong, Weihan Wang, Ming Ding, Wenmeng Yu, Qingsong Lv, Yan Wang, Yean Cheng, Shiyu Huang, Junhui Ji, Zhao Xue, et~al.
\newblock Cogvlm2: Visual language models for image and video understanding.
\newblock {\em arXiv preprint arXiv:2408.16500}, 2024.

\bibitem{liu2024oryx}
Zuyan Liu, Yuhao Dong, Ziwei Liu, Winston Hu, Jiwen Lu, and Yongming Rao.
\newblock Oryx mllm: On-demand spatial-temporal understanding at arbitrary resolution.
\newblock {\em arXiv preprint arXiv:2409.12961}, 2024.

\bibitem{zhang2024STGT}
Shi-Xue Zhang, Hongfa Wang, Xiaobin Zhu, Weibo Gu, Tianjin Zhang, Chun Yang, Wei Liu, and Xu-Cheng Yin.
\newblock Video-language alignment pre-training via spatio-temporal graph transformer.
\newblock {\em arXiv e-prints}, pages arXiv--2407, 2024.

\bibitem{li2024aria}
Dongxu Li, Yudong Liu, Haoning Wu, Yue Wang, Zhiqi Shen, Bowen Qu, Xinyao Niu, Guoyin Wang, Bei Chen, and Junnan Li.
\newblock Aria: An open multimodal native mixture-of-experts model.
\newblock {\em arXiv preprint arXiv:2410.05993}, 2024.

\bibitem{SORA}
Tim Brooks, Bill Peebles, Connor Holmes, Will DePue, Yufei Guo, Li~Jing, David Schnurr, Joe Taylor, Troy Luhman, Eric Luhman, et~al.
\newblock Video generation models as world simulators, 2024.

\bibitem{vidu}
Fan Bao, Chendong Xiang, Gang Yue, Guande He, Hongzhou Zhu, Kaiwen Zheng, Min Zhao, Shilong Liu, Yaole Wang, and Jun Zhu.
\newblock Vidu: a highly consistent, dynamic and skilled text-to-video generator with diffusion models.
\newblock {\em arXiv preprint arXiv:2405.04233}, 2024.

\bibitem{kelin}
Ye~Tian, Ling Yang, Haotian Yang, Yuan Gao, Yufan Deng, Jingmin Chen, Xintao Wang, Zhaochen Yu, Xin Tao, Pengfei Wan, et~al.
\newblock Videotetris: Towards compositional text-to-video generation.
\newblock {\em arXiv preprint arXiv:2406.04277}, 2024.

\bibitem{hunyuanvideo}
Weijie Kong, Qi~Tian, Zijian Zhang, Rox Min, Zuozhuo Dai, Jin Zhou, Jiangfeng Xiong, Xin Li, Bo~Wu, Jianwei Zhang, et~al.
\newblock Hunyuanvideo: A systematic framework for large video generative models.
\newblock {\em arXiv preprint arXiv:2412.03603}, 2024.

\bibitem{zhang2024video}
Yuanhan Zhang, Jinming Wu, Wei Li, Bo~Li, Zejun Ma, Ziwei Liu, and Chunyuan Li.
\newblock Video instruction tuning with synthetic data.
\newblock {\em arXiv preprint arXiv:2410.02713}, 2024.

\bibitem{xu2024pllava}
Lin Xu, Yilin Zhao, Daquan Zhou, Zhijie Lin, See~Kiong Ng, and Jiashi Feng.
\newblock Pllava: Parameter-free llava extension from images to videos for video dense captioning.
\newblock {\em arXiv preprint arXiv:2404.16994}, 2024.

\bibitem{wang2024qwen2}
Peng Wang, Shuai Bai, Sinan Tan, Shijie Wang, Zhihao Fan, Jinze Bai, Keqin Chen, Xuejing Liu, Jialin Wang, Wenbin Ge, et~al.
\newblock Qwen2-vl: Enhancing vision-language model's perception of the world at any resolution.
\newblock {\em arXiv preprint arXiv:2409.12191}, 2024.

\bibitem{li2024llava}
Bo~Li, Yuanhan Zhang, Dong Guo, Renrui Zhang, Feng Li, Hao Zhang, Kaichen Zhang, Peiyuan Zhang, Yanwei Li, Ziwei Liu, et~al.
\newblock Llava-onevision: Easy visual task transfer.
\newblock {\em arXiv preprint arXiv:2408.03326}, 2024.

\bibitem{li2024mvbench}
Kunchang Li, Yali Wang, Yinan He, Yizhuo Li, Yi~Wang, Yi~Liu, Zun Wang, Jilan Xu, Guo Chen, Ping Luo, et~al.
\newblock Mvbench: A comprehensive multi-modal video understanding benchmark.
\newblock In {\em Proceedings of the IEEE/CVF Conference on Computer Vision and Pattern Recognition}, pages 22195--22206, 2024.

\bibitem{li2024videovista}
Yunxin Li, Xinyu Chen, Baotian Hu, Longyue Wang, Haoyuan Shi, and Min Zhang.
\newblock Videovista: A versatile benchmark for video understanding and reasoning.
\newblock {\em arXiv preprint arXiv:2406.11303}, 2024.

\bibitem{wang2024lvbench}
Weihan Wang, Zehai He, Wenyi Hong, Yean Cheng, Xiaohan Zhang, Ji~Qi, Shiyu Huang, Bin Xu, Yuxiao Dong, Ming Ding, et~al.
\newblock Lvbench: An extreme long video understanding benchmark.
\newblock {\em arXiv preprint arXiv:2406.08035}, 2024.

\bibitem{fu2024video}
Chaoyou Fu, Yuhan Dai, Yongdong Luo, Lei Li, Shuhuai Ren, Renrui Zhang, Zihan Wang, Chenyu Zhou, Yunhang Shen, Mengdan Zhang, et~al.
\newblock Video-mme: The first-ever comprehensive evaluation benchmark of multi-modal llms in video analysis.
\newblock {\em arXiv preprint arXiv:2405.21075}, 2024.

\bibitem{zhou2024mlvu}
Junjie Zhou, Yan Shu, Bo~Zhao, Boya Wu, Shitao Xiao, Xi~Yang, Yongping Xiong, Bo~Zhang, Tiejun Huang, and Zheng Liu.
\newblock Mlvu: A comprehensive benchmark for multi-task long video understanding.
\newblock {\em arXiv preprint arXiv:2406.04264}, 2024.

\bibitem{wu2024longvideobench}
Haoning Wu, Dongxu Li, Bei Chen, and Junnan Li.
\newblock Longvideobench: A benchmark for long-context interleaved video-language understanding.
\newblock {\em arXiv preprint arXiv:2407.15754}, 2024.

\bibitem{meteor}
Satanjeev Banerjee and Alon Lavie.
\newblock Meteor: An automatic metric for mt evaluation with improved correlation with human judgments.
\newblock In {\em Proceedings of the acl workshop on intrinsic and extrinsic evaluation measures for machine translation and/or summarization}, pages 65--72, 2005.

\bibitem{bleu}
Kishore Papineni, Salim Roukos, Todd Ward, and Wei-Jing Zhu.
\newblock Bleu: a method for automatic evaluation of machine translation.
\newblock In {\em Proceedings of the 40th annual meeting of the Association for Computational Linguistics}, pages 311--318, 2002.

\bibitem{Spice}
Peter Anderson, Basura Fernando, Mark Johnson, and Stephen Gould.
\newblock Spice: Semantic propositional image caption evaluation.
\newblock In {\em Computer Vision--ECCV 2016: 14th European Conference, Amsterdam, The Netherlands, October 11-14, 2016, Proceedings, Part V 14}, pages 382--398. Springer, 2016.

\bibitem{cider}
Ramakrishna Vedantam, C~Lawrence~Zitnick, and Devi Parikh.
\newblock Cider: Consensus-based image description evaluation.
\newblock In {\em Proceedings of the IEEE conference on computer vision and pattern recognition}, pages 4566--4575, 2015.

\bibitem{infometic}
Anwen Hu, Shizhe Chen, Liang Zhang, and Qin Jin.
\newblock Infometic: An informative metric for reference-free image caption evaluation.
\newblock {\em arXiv preprint arXiv:2305.06002}, 2023.

\bibitem{Clipscore}
Jack Hessel, Ari Holtzman, Maxwell Forbes, Ronan~Le Bras, and Yejin Choi.
\newblock Clipscore: A reference-free evaluation metric for image captioning.
\newblock {\em arXiv preprint arXiv:2104.08718}, 2021.

\bibitem{Tiger}
Ming Jiang, Qiuyuan Huang, Lei Zhang, Xin Wang, Pengchuan Zhang, Zhe Gan, Jana Diesner, and Jianfeng Gao.
\newblock Tiger: Text-to-image grounding for image caption evaluation.
\newblock {\em arXiv preprint arXiv:1909.02050}, 2019.

\bibitem{Faier}
Sijin Wang, Ziwei Yao, Ruiping Wang, Zhongqin Wu, and Xilin Chen.
\newblock Faier: Fidelity and adequacy ensured image caption evaluation.
\newblock In {\em Proceedings of the IEEE/CVF Conference on Computer Vision and Pattern Recognition}, pages 14050--14059, 2021.

\bibitem{QACE}
Hwanhee Lee, Thomas Scialom, Seunghyun Yoon, Franck Dernoncourt, and Kyomin Jung.
\newblock Qace: Asking questions to evaluate an image caption.
\newblock {\em arXiv preprint arXiv:2108.12560}, 2021.

\bibitem{cho2023davidsonian}
Jaemin Cho, Yushi Hu, Roopal Garg, Peter Anderson, Ranjay Krishna, Jason Baldridge, Mohit Bansal, Jordi Pont-Tuset, and Su~Wang.
\newblock Davidsonian scene graph: Improving reliability in fine-grained evaluation for text-image generation.
\newblock {\em arXiv preprint arXiv:2310.18235}, 2023.

\bibitem{Ella}
Xiwei Hu, Rui Wang, Yixiao Fang, Bin Fu, Pei Cheng, and Gang Yu.
\newblock Ella: Equip diffusion models with llm for enhanced semantic alignment.
\newblock {\em arXiv preprint arXiv:2403.05135}, 2024.

\bibitem{playgroundv3}
Bingchen Liu, Ehsan Akhgari, Alexander Visheratin, Aleks Kamko, Linmiao Xu, Shivam Shrirao, Chase Lambert, Joao Souza, Suhail Doshi, and Daiqing Li.
\newblock Playground v3: Improving text-to-image alignment with deep-fusion large language models.
\newblock {\em arXiv preprint arXiv:2409.10695}, 2024.

\bibitem{clip}
Alec Radford, Jong~Wook Kim, Chris Hallacy, Aditya Ramesh, Gabriel Goh, Sandhini Agarwal, Girish Sastry, Amanda Askell, Pamela Mishkin, Jack Clark, et~al.
\newblock Learning transferable visual models from natural language supervision.
\newblock In {\em International conference on machine learning}, pages 8748--8763. PMLR, 2021.

\bibitem{ChenSMDCJF0RYT24}
Tsai{-}Shien Chen, Aliaksandr Siarohin, Willi Menapace, Ekaterina Deyneka, Hsiang{-}wei Chao, Byung~Eun Jeon, Yuwei Fang, Hsin{-}Ying Lee, Jian Ren, Ming{-}Hsuan Yang, and Sergey Tulyakov.
\newblock Panda-70m: Captioning 70m videos with multiple cross-modality teachers.
\newblock In {\em {IEEE/CVF} Conference on Computer Vision and Pattern Recognition, {CVPR} 2024, Seattle, WA, USA, June 16-22, 2024}, pages 13320--13331. {IEEE}, 2024.

\bibitem{grauman2022ego4d}
Kristen Grauman, Andrew Westbury, Eugene Byrne, Zachary Chavis, Antonino Furnari, Rohit Girdhar, Jackson Hamburger, Hao Jiang, Miao Liu, Xingyu Liu, et~al.
\newblock Ego4d: Around the world in 3,000 hours of egocentric video.
\newblock In {\em Proceedings of the IEEE/CVF conference on computer vision and pattern recognition}, pages 18995--19012, 2022.

\bibitem{yu2020bdd100k}
Fisher Yu, Haofeng Chen, Xin Wang, Wenqi Xian, Yingying Chen, Fangchen Liu, Vashisht Madhavan, and Trevor Darrell.
\newblock Bdd100k: A diverse driving dataset for heterogeneous multitask learning.
\newblock In {\em Proceedings of the IEEE/CVF conference on computer vision and pattern recognition}, pages 2636--2645, 2020.

\bibitem{chensharegpt4video}
Lin Chen, Xilin Wei, Jinsong Li, Xiaoyi Dong, Pan Zhang, Yuhang Zang, Zehui Chen, Haodong Duan, Bin Lin, Zhenyu Tang, et~al.
\newblock Sharegpt4video: Improving video understanding and generation with better captions.
\newblock In {\em The Thirty-eight Conference on Neural Information Processing Systems Datasets and Benchmarks Track}.

\bibitem{tan2024vidgen}
Zhiyu Tan, Xiaomeng Yang, Luozheng Qin, and Hao Li.
\newblock Vidgen-1m: A large-scale dataset for text-to-video generation.
\newblock {\em arXiv preprint arXiv:2408.02629}, 2024.

\bibitem{yuan2024chronomagic}
Shenghai Yuan, Jinfa Huang, Yongqi Xu, Yaoyang Liu, Shaofeng Zhang, Yujun Shi, Ruijie Zhu, Xinhua Cheng, Jiebo Luo, and Li~Yuan.
\newblock Chronomagic-bench: A benchmark for metamorphic evaluation of text-to-time-lapse video generation.
\newblock {\em arXiv preprint arXiv:2406.18522}, 2024.

\bibitem{Farre2024FineVideo}
Miquel Farré, Andi Marafioti, Lewis Tunstall, Leandro Von~Werra, and Thomas Wolf.
\newblock Finevideo.
\newblock 2024.

\bibitem{wang2024lift}
Yibin Wang, Zhiyu Tan, Junyan Wang, Xiaomeng Yang, Cheng Jin, and Hao Li.
\newblock Lift: Leveraging human feedback for text-to-video model alignment.
\newblock {\em arXiv preprint arXiv:2412.04814}, 2024.

\bibitem{bai2025qwen2}
Shuai Bai, Keqin Chen, Xuejing Liu, Jialin Wang, Wenbin Ge, Sibo Song, Kai Dang, Peng Wang, Shijie Wang, Jun Tang, et~al.
\newblock Qwen2. 5-vl technical report.
\newblock {\em arXiv preprint arXiv:2502.13923}, 2025.

\bibitem{zhang2024videoinstructiontuningsynthetic}
Yuanhan Zhang, Jinming Wu, Wei Li, Bo~Li, Zejun Ma, Ziwei Liu, and Chunyuan Li.
\newblock Video instruction tuning with synthetic data, 2024.

\bibitem{liu2024nvila}
Zhijian Liu, Ligeng Zhu, Baifeng Shi, Zhuoyang Zhang, Yuming Lou, Shang Yang, Haocheng Xi, Shiyi Cao, Yuxian Gu, Dacheng Li, et~al.
\newblock Nvila: Efficient frontier visual language models.
\newblock {\em arXiv preprint arXiv:2412.04468}, 2024.

\bibitem{zhang2025videollama}
Boqiang Zhang, Kehan Li, Zesen Cheng, Zhiqiang Hu, Yuqian Yuan, Guanzheng Chen, Sicong Leng, Yuming Jiang, Hang Zhang, Xin Li, et~al.
\newblock Videollama 3: Frontier multimodal foundation models for image and video understanding.
\newblock {\em arXiv preprint arXiv:2501.13106}, 2025.

\end{thebibliography}

\clearpage
\appendix
\section{Technical Appendices and Supplementary Material}
In this section, we present the automated evaluation results, employing GPT-4.1 as the evaluation expert. This analysis aims to provide researchers with deeper insights into the characteristics of our dataset, as well as the variations in outcomes attributable to different evaluation experts. Additionally, we include the original caption outputs and corresponding evaluation results produced by various models for comprehensive comparison.

\subsection{More Experimental Results With GPT-4.1}\label{exp_app}

\begin{table*}[htbp]
\renewcommand{\arraystretch}{0.6}
\caption{The accuracy (AR), inconsistency rate (IR), and coverage rate (CR) of VLM methods on all dimensions, where GPT-4.1 as a TextQA expert. The symbol ``$\uparrow$"  indicates that the larger the value, the better; The symbol ``$\downarrow$" indicates that the smaller the value, the better.}
\label{tab:results_2}
\resizebox{\textwidth}{!}{%
\begin{tabular}{@{}ll|cccccccccccc@{}}
\toprule
 &\multirow{3}{*}{Methods}& \multicolumn{7}{c|}{Content \& Entity}& \multicolumn{4}{c}{Color \&  Light}\\
 \cline{3-13} 
 &  & \begin{tabular}[c]{@{}c@{}}\small Proper\\ \small Noun\end{tabular} & \begin{tabular}[c]{@{}c@{}}Action\end{tabular} & \begin{tabular}[c]{@{}c@{}}Relation
\end{tabular} & \begin{tabular}[c]{@{}c@{}}Count
\end{tabular} & Entity  & \begin{tabular}[c]{@{}c@{}}\small Entity\\ \small Size\end{tabular} & \multicolumn{1}{c|}{\begin{tabular}[c]{@{}c@{}}\small Entity\\ \small Shape\end{tabular} }
&\begin{tabular}[c]{@{}c@{}}Lighting
\end{tabular} & \begin{tabular}[c]{@{}c@{}}\small Color \\ \small Palette \end{tabular} & \begin{tabular}[c]{@{}c@{}}\small Color\\ \small Grading
\end{tabular} & \begin{tabular}[c]{@{}c@{}}Color
\end{tabular}  \\ \midrule
\multirow{8}{*}{\rotatebox{90}{AR}$\Bigg\uparrow$} 
& LLaVA-Video-7B & 40.73 & 57.08 & 51.67 & 54.88 & 34.02 & 66.2 & \multicolumn{1}{c|}{26.62} & 57.59 & 61.49 & 23.78 & 55.97 \\
& Qwen2VL-7B & 39.17 & 56.32 & 52.87 & 57.17 & 32.27 & 64.2 & \multicolumn{1}{c|}{25.61} & 69.93 & 55.57 & 30.62 & \multicolumn{1}{c|}{50.75}  \\ 
& VideoLLaMA3-7B & 38.92 & 58.51 & 52.56 & 56.27 & 33.48 & 64.28 & \multicolumn{1}{c|}{25.56} & 60.59 & 59.03 & 25.24 & \multicolumn{1}{c|}{54.12}\\
& NVILA-8B & 40.03 & 44.2 & 48.04 & 57.9 & 37.3 & 63.72 & \multicolumn{1}{c|}{29.32} & 72.63 & 66.01 & 42.37 & \multicolumn{1}{c|}{59.35} \\ 
& InternVL2.5-8B & 38.63 & 53.86 & 48.71 & 59.64 & 36.78 & 60.91 & \multicolumn{1}{c|}{29.44} & 81.27 & 75.35 & 60.36 & \multicolumn{1}{c|}{55.82} \\
& Qwen2.5VL-7B & 46.95 & 63.86 & 57.72 & 62.39 & 41.75 & 69.35 & \multicolumn{1}{c|}{34.89} & 83.6 & 73.9 & 51.32 & \multicolumn{1}{c|}{59.8}  \\ 
& Qwen2.5-VL-72B & 48.76 & 67.28 & 63.03 & 67.33 & 45.37 & 72.51 & \multicolumn{1}{c|}{37.76} & 88.42 & 75.28 & 61.31 & \multicolumn{1}{c|}{62.69}  \\ 
& GPT-4o  & 51.11 & 69.48 & 66.34 & 70.08 & 51.38 & 77.5 & \multicolumn{1}{c|}{47.89} & 87.96 & 82.47 & 68.17 & \multicolumn{1}{c|}{67.07} \\ 
& Gemini2.5-Pro-Flash & 60.21 & 76.23 & 73.54 & 79.59 & 63.97 & 85.25 & \multicolumn{1}{c|}{59.55} & \textbf{90.05} & 86.57 & \textbf{69.06} & \multicolumn{1}{c|}{79.36}\\ 
& Gemini-2.5-Pro-Preview & \textbf{61.41} & \textbf{78.13} & \textbf{75.19} & \textbf{80.54} & \textbf{64.4} & \textbf{86.52} & \multicolumn{1}{c|}{\textbf{62.21}} & 88.9 & \textbf{88.06} & 66.95 & \multicolumn{1}{c|}{\textbf{79.96}} \\ 

\midrule
\multirow{8}{*}{\rotatebox{90}{IR}$\Bigg\downarrow$} 
& LLaVA-Video-7B & 13.09 & 15.13 & 11.29 & 13.14 & 3.76 & 8.62 & \multicolumn{1}{c|}{8.45} & \textbf{5.62} & 11.05 & \textbf{4.12} & \multicolumn{1}{c|}{13.23} \\
& Qwen2VL-7B & 8.38 & 13.84 & 10.27 & 11.67 & 3.77 & 8.64 & \multicolumn{1}{c|}{\textbf{6.05}} & 7.26 & 12.88 & 5.67 & \multicolumn{1}{c|}{13.44} \\
& VideoLLaMA3-7B & 9.91 & 17.88 & 11.99 & 12.59 & 3.46 & 8.41 & \multicolumn{1}{c|}{7.56} & 6.19 & 11.49 & 4.39 & \multicolumn{1}{c|}{12.77}  \\
& NVILA-8B & 11.88 & 20.21 & 14.96 & 15.45 & 4.48 & 12.42 & \multicolumn{1}{c|}{9.61} & 8.08 & 15.72 & 6.52 & \multicolumn{1}{c|}{17.75}  \\
& InternVL2.5-8B & 14.57 & 19.0 & 16.43 & 14.18 & 5.5 & 14.0 & \multicolumn{1}{c|}{10.23} & 8.06 & 15.11 & 8.57 & \multicolumn{1}{c|}{16.01} \\
& Qwen2.5VL-7B & 8.65 & 14.66 & 12.39 & 11.35 & \textbf{3.16} & 9.25 & \multicolumn{1}{c|}{8.38} & 6.9 & 11.5 & 5.52 & \multicolumn{1}{c|}{13.87} \\

& Qwen2.5-VL-72B & 7.86 & 15.0 & 11.58 & 11.66 & 5.02 & 8.96 & \multicolumn{1}{c|}{8.28} & 6.89 & 13.44 & 5.3 & \multicolumn{1}{c|}{15.23} \\ 
& GPT-4o & 7.53 & \textbf{11.26} & \textbf{9.56} & 10.45 & 3.94 & 6.5 & \multicolumn{1}{c|}{8.48} & 6.34 & 11.69 & 7.32 & \multicolumn{1}{c|}{13.15} \\ 
& Gemini2.5-Pro-Flash & 6.76 & 12.59 & 9.57 & 9.81 & 5.23 & \textbf{5.36} & \multicolumn{1}{c|}{8.43} & 6.43 & 11.32 & 6.98 & \multicolumn{1}{c|}{\textbf{11.19}} \\
& Gemini-2.5-Pro-Preview & \textbf{6.58}& 13.06 & 9.77 & \textbf{9.06} & 4.29 & 5.88 & \multicolumn{1}{c|}{8.64} & 7.08 & \textbf{10.16} & 5.76 & \multicolumn{1}{c|}{12.18}\\ 

 \midrule
\multirow{8}{*}{\rotatebox{90}{CR}$\Bigg\uparrow$} 
& LLaVA-Video-7B & 46.87 & 67.25 & 58.25 & 63.17 & 35.35 & 72.44 & \multicolumn{1}{c|}{29.07} & 61.03 & 69.13 & 24.81 & \multicolumn{1}{c|}{64.51}   \\
& Qwen2VL-7B & 42.75 & 65.36 & 58.92 & 64.73 & 33.54 & 70.27 & \multicolumn{1}{c|}{27.26} & 75.41 & 63.79 & 32.46 & \multicolumn{1}{c|}{58.63}  \\ 
& VideoLLaMA3-7B & 43.2 & 71.25 & 59.72 & 64.38 & 34.68 & 70.19 & \multicolumn{1}{c|}{27.65} & 64.59 & 66.7 & 26.4 & \multicolumn{1}{c|}{62.05}  \\
& NVILA-8B & 45.43 & 55.4 & 56.49 & 68.48 & 39.05 & 72.76 & \multicolumn{1}{c|}{32.43} & 79.01 & 78.33 & 45.32 & \multicolumn{1}{c|}{72.16}  \\
& InternVL2.5-8B & 45.22 & 66.49 & 58.29 & 69.49 & 38.92 & 70.82 & \multicolumn{1}{c|}{32.8} & 88.4 & 88.75 & 66.02 & \multicolumn{1}{c|}{66.46}   \\
& Qwen2.5VL-7B & 51.4 & 74.83 & 65.88 & 70.37 & 43.11 & 76.42 & \multicolumn{1}{c|}{38.08} & 89.79 & 83.5 & 54.32 & \multicolumn{1}{c|}{69.43} \\ 
& Qwen2.5-VL-72B & 52.92 & 79.16 & 71.28 & 76.21 & 47.76 & 79.64 & \multicolumn{1}{c|}{41.17} & 94.96 & 86.97 & 64.74 & \multicolumn{1}{c|}{73.96}  \\ 

& GPT-4o & 55.27 & 78.3 & 73.36 & 78.26 & 53.48 & 82.89 & \multicolumn{1}{c|}{52.33} & 93.92 & 93.39 & 73.56 & \multicolumn{1}{c|}{77.23}\\ 
& Gemini2.5-Pro-Flash & 64.58 & 87.22 & 81.32 & 88.25 & \textbf{67.5} & 90.07 & \multicolumn{1}{c|}{65.03} & \textbf{96.24} & 97.62 & \textbf{74.24} & \multicolumn{1}{c|}{89.36}\\ 
& Gemini-2.5-Pro-Preview & \textbf{65.73} & \textbf{89.86} & \textbf{83.33} & \textbf{88.56} & 67.28 & \textbf{91.93} & \multicolumn{1}{c|}{\textbf{68.1}} & 95.68 & \textbf{98.02} & 71.04 & \multicolumn{1}{c|}{\textbf{91.04}} \\

 \bottomrule
 \toprule
 &\multirow{3}{*}{Methods}& \multicolumn{5}{c|}{Visuals \& Composition}& \multicolumn{5}{c|}{Cinematography  \& Environment}\\
 \cline{3-13} 
 &  & Position & \begin{tabular}[c]{@{}c@{}}\small Relative\\ \small Position\end{tabular} & Background &Text & \multicolumn{1}{c|}{Blur} & Style & \begin{tabular}[c]{@{}c@{}}\small Camera \\ \small Movement\end{tabular} & \begin{tabular}[c]{@{}c@{}}Shot Type\end{tabular} & \begin{tabular}[c]{@{}c@{}}Emotion\end{tabular} & \multicolumn{1}{c|}{Atmosphere} & \color{red}\textbf{ALL}\\ \midrule
\multirow{8}{*}{\rotatebox{90}{AR}$\Bigg\uparrow$} 

& LLaVA-Video-7B & 34.26 & 32.96 & 69.28 & 43.75 & \multicolumn{1}{c|}{13.49} & 66.6 & 23.68 & 38.13 & 48.53 & \multicolumn{1}{c|}{82.04} & 47.74 \\
& Qwen2VL-7B & 35.34 & 32.4 & 68.69 & 34.75 & \multicolumn{1}{c|}{15.17} & 71.53 & 16.26 & 32.7 & 55.24 & \multicolumn{1}{c|}{83.56} & 47.45\\ 
& VideoLLaMA3-7B & 35.41 & 32.54 & 65.35 & 38.11 & \multicolumn{1}{c|}{16.99} & 71.66 & 39.32 & 39.53 & 53.27 & \multicolumn{1}{c|}{80.44} &48.45  \\
& NVILA-8B & 40.69 & 33.36 & 69.7 & 38.04 & \multicolumn{1}{c|}{26.42} & 74.11 & 16.97 & 39.4 & 53.46 & \multicolumn{1}{c|}{81.02} &49.70 \\
& InternVL2.5-8B & 39.08 & 32.59 & 70.89 & 37.31 & \multicolumn{1}{c|}{36.83} & 86.59 & 51.3 & 50.53 & 58.75 & \multicolumn{1}{c|}{86.57} &54.85  \\
 & Qwen2.5VL-7B & 45.0 & 38.58 & 77.18 & 44.96 & \multicolumn{1}{c|}{30.71} & 88.55 & 51.92 & 50.9 & 63.5 & \multicolumn{1}{c|}{89.51} &58.74 \\ 
& Qwen2.5-VL-72B & 49.83 & 43.01 & 78.99 & 45.55 & \multicolumn{1}{c|}{36.31} & 92.0 & 66.47 & 59.37 & 67.78 & \multicolumn{1}{c|}{89.91} &63.08 \\
& GPT-4o & 57.24 & 49.0 & 83.14 & 46.79 & \multicolumn{1}{c|}{48.0} & 92.05 & 55.57 & 62.97 & 62.45 & \multicolumn{1}{c|}{\textbf{90.82}} &66.49\\
& Gemini2.5-Pro-Flash & 70.36 & 63.56 & 87.64 & \textbf{63.7} & \multicolumn{1}{c|}{67.33} & \textbf{96.13} & 70.76 & 74.29 & 73.01 & \multicolumn{1}{c|}{86.22} &75.56 \\
& Gemini-2.5-Pro-Preview & \textbf{70.63} & \textbf{66.11} & \textbf{87.97} & 63.5 & \multicolumn{1}{c|}{\textbf{68.75}} & 95.76 & \textbf{74.89} & \textbf{75.72} & \textbf{77.05} & \multicolumn{1}{c|}{88.34} &\color{red}\textbf{76.73}\\

 \midrule
\multirow{8}{*}{\rotatebox{90}{IR}$\Bigg\downarrow$} 
& LLaVA-Video-7B & \textbf{10.46} & 11.83 & 7.5 & 17.92 & \multicolumn{1}{c|}{\textbf{10.11}} & \textbf{1.64} & 24.13 & \textbf{12.18} & \textbf{3.67} & \multicolumn{1}{c|}{\textbf{3.71}} & 10.05\\
& Qwen2VL-7B & 11.18 & \textbf{10.39} & 8.8 & 13.47 & \multicolumn{1}{c|}{14.47} & 2.72 & 19.21 & 13.17 & 5.16 & \multicolumn{1}{c|}{6.25} & \color{red}\textbf{9.65} \\
& VideoLLaMA3-7B & 12.26 & 12.21 & 7.94 & 14.86 & \multicolumn{1}{c|}{12.18} & 2.22 & 23.44 & 14.68 & 3.74 & \multicolumn{1}{c|}{4.52} &10.50\\
& NVILA-8B & 16.22 & 16.46 & 10.26 & 17.93 & \multicolumn{1}{c|}{18.62} & 4.51 & 23.6 & 22.5 & 6.31 & \multicolumn{1}{c|}{6.05} &13.12\\
& InternVL2.5-8B & 15.32 & 14.72 & 8.74 & 19.13 & \multicolumn{1}{c|}{21.88} & 4.89 & 26.24 & 20.04 & 6.11 & \multicolumn{1}{c|}{5.57} &13.41 \\
& Qwen2.5VL-7B & 12.8 & 13.45 & 8.37 & 12.25 & \multicolumn{1}{c|}{19.8} & 2.72 & 22.77 & 16.43 & 5.47 & \multicolumn{1}{c|}{5.09} & 10.62\\

& Qwen2.5-VL-72B & 13.34 & 13.5 & 7.98 & 12.56 & \multicolumn{1}{c|}{20.73} & 3.36 & 20.99 & 16.97 & 5.56 & \multicolumn{1}{c|}{5.14} &10.98\\
& GPT-4o & 12.42 & 12.35 & 6.23 & 12.9 & \multicolumn{1}{c|}{22.32} & 3.64 & 21.35 & 19.02 & 6.91 & \multicolumn{1}{c|}{5.06} &10.24\\

&  Gemini2.5-Pro-Flash & 14.68 & 15.16 & \textbf{5.05} & 11.87 & \multicolumn{1}{c|}{23.45} & 2.25 & 20.57 & 20.34 & 5.12 & \multicolumn{1}{c|}{4.79} &10.49\\

&Gemini-2.5-Pro-Preview & 15.65 & 13.98 & 5.12 & \textbf{10.92} & \multicolumn{1}{c|}{22.58} & 2.33 & \textbf{18.68} & 19.17 & 4.7 & \multicolumn{1}{c|}{4.7} &10.28\\

 \midrule
\multirow{8}{*}{\rotatebox{90}{CR}$\Bigg\uparrow$} 
& LLaVA-Video-7B & 38.26 & 37.38 & 74.9 & 53.3 & \multicolumn{1}{c|}{15.01} & 67.71 & 31.22 & 43.42 & 50.38 & \multicolumn{1}{c|}{85.2} & 53.08\\
& Qwen2VL-7B & 39.79 & 36.16 & 75.32 & 40.16 & \multicolumn{1}{c|}{17.74} & 73.53 & 20.13 & 37.66 & 58.25 & \multicolumn{1}{c|}{89.14}  & 52.52\\ 
& VideoLLaMA3-7B & 40.36 & 37.07 & 70.98 & 44.76 & \multicolumn{1}{c|}{19.35} & 73.29 & 51.37 & 46.33 & 55.34 & \multicolumn{1}{c|}{84.25}  & 54.13\\
& NVILA-8B & 48.57 & 39.93 & 77.67 & 46.35 & \multicolumn{1}{c|}{32.46} & 77.61 & 22.22 & 50.83 & 57.06 & \multicolumn{1}{c|}{86.24}  & 57.21\\
& InternVL2.5-8B & 46.15 & 38.22 & 77.69 & 46.14 & \multicolumn{1}{c|}{47.14} & 91.05 & 69.55 & 63.19 & 62.57 & \multicolumn{1}{c|}{91.68} & 63.34\\
& Qwen2.5VL-7B & 51.6 & 44.58 & 84.23 & 51.24 & \multicolumn{1}{c|}{38.3} & 91.03 & 67.23 & 60.91 & 67.17 & \multicolumn{1}{c|}{94.31} &65.75 \\
& Qwen2.5-VL-72B & 57.5 & 49.72 & 85.84 & 52.1 & \multicolumn{1}{c|}{45.81} & 95.2 & 84.14 & 71.51 & 71.78 & \multicolumn{1}{c|}{94.78} &70.86\\
& GPT-4o & 65.36 & 55.91 & 88.67 & 53.72 & \multicolumn{1}{c|}{61.8} & 95.52 & 70.65 & 77.77 & 67.09 & \multicolumn{1}{c|}{\textbf{95.66}} &74.07\\
& Gemini2.5-Pro-Flash & 82.47 & 74.92 & 92.3 & \textbf{72.27} & \multicolumn{1}{c|}{87.96} & \textbf{98.35} & 89.09 & 93.26 & 76.95 & \multicolumn{1}{c|}{90.55} &84.42\\
& Gemini-2.5-Pro-Preview & \textbf{83.74} & \textbf{76.86} & \textbf{92.72} & 71.28 & \multicolumn{1}{c|}{\textbf{88.8}} & 98.04 & \textbf{92.09} & \textbf{93.68} & \textbf{80.85} & \multicolumn{1}{c|}{92.7}&\color{red}\textbf{85.52} \\

\bottomrule
\end{tabular}%
}
\end{table*}

\begin{figure*}[tp]
	\begin{minipage}[t]{1.0\linewidth}
        \centering
	\includegraphics[width=1.0\linewidth]{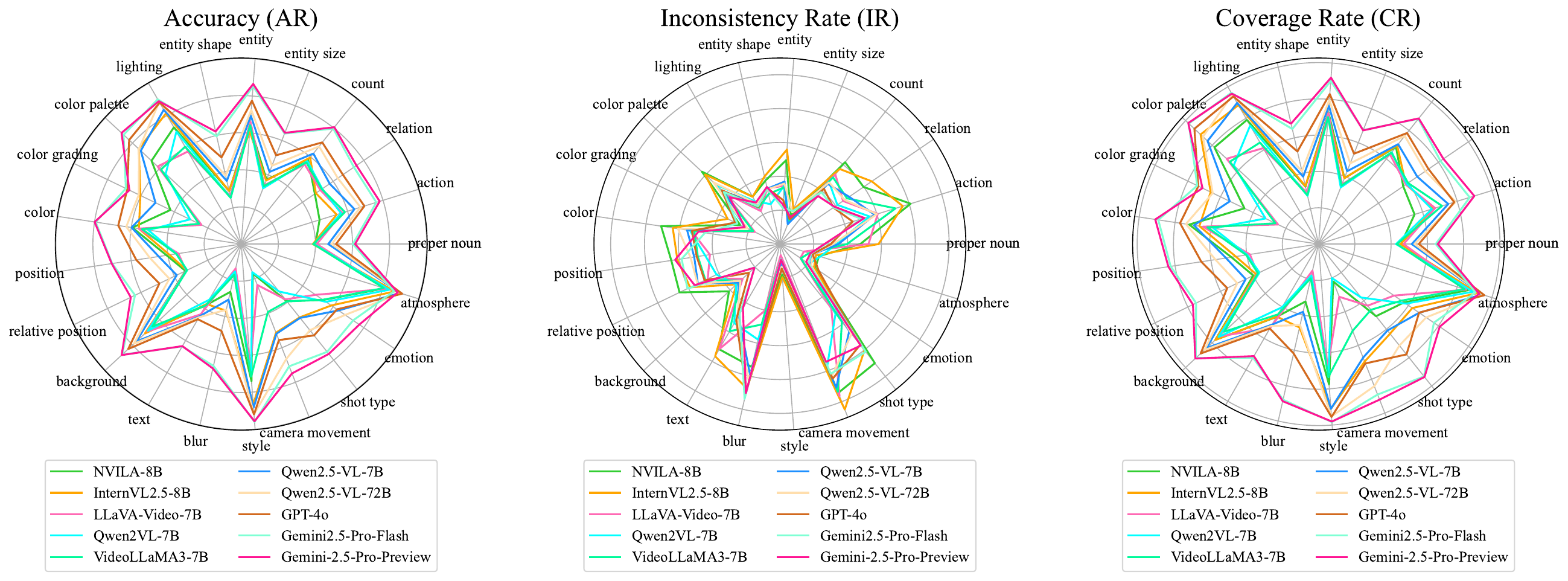}
	\caption{Results of GPT-4.1 captioning evaluation, organized by category.}
	\label{fig:fig9}
	\end{minipage}%
\end{figure*}

\begin{figure}[thbp]
	\begin{minipage}[t]{1.0\linewidth}
        \centering
	\includegraphics[width=0.9\linewidth]{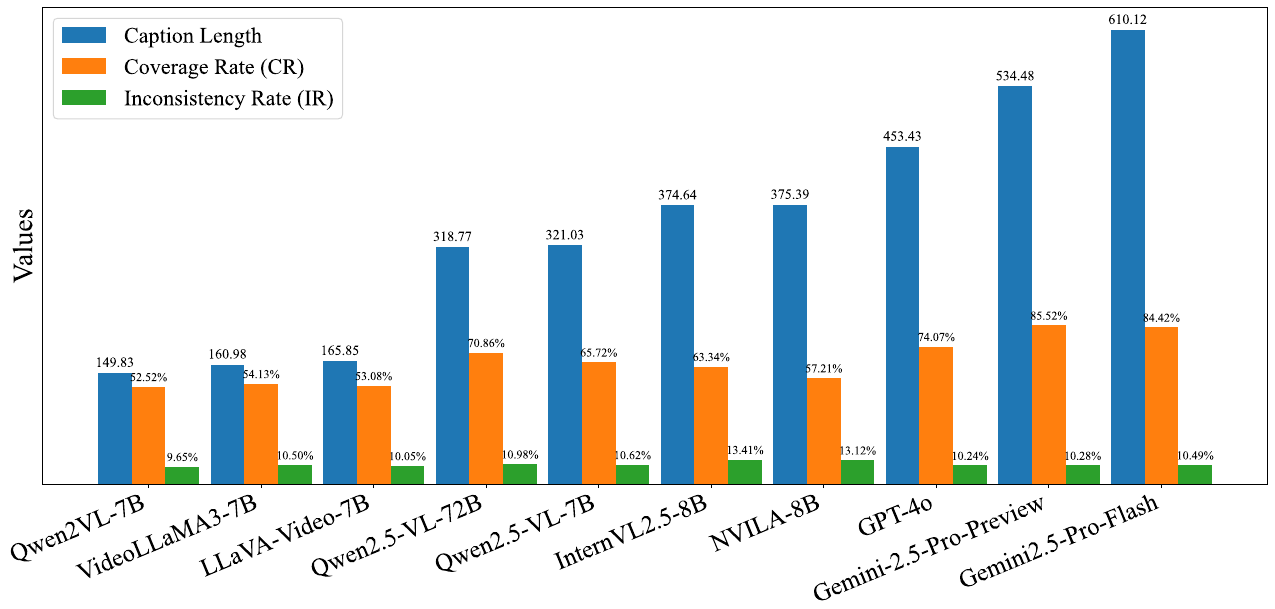}
    \vspace{-0.5em}
	\caption{The relationship between CR, IR and caption length, where GPT-4.1 as a TextQA expert. Caption length is the number of words in the caption, not counting special symbols.}
	\label{fig:fig10}
	\end{minipage}%
\end{figure}

\begin{figure}[thbp]
	\begin{minipage}[t]{1.0\linewidth}
        \centering
	\includegraphics[width=0.9\linewidth]{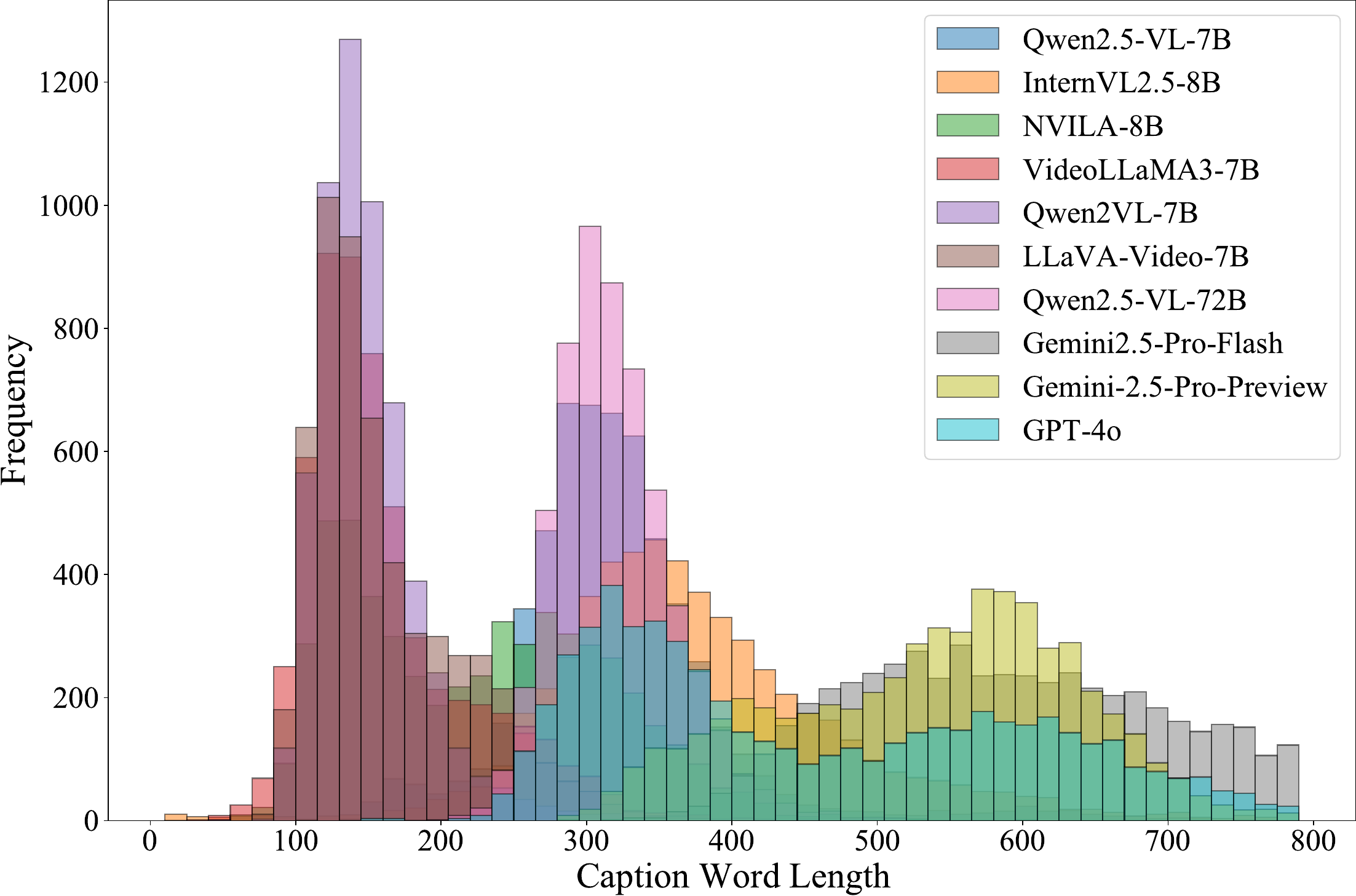}
	\caption{Video caption word length distribution.}
	\label{fig:fig11}
	\end{minipage}%
\end{figure}

Table~\ref{tab:results_2} and Fig.~\ref{fig:fig9} presents the accuracy rate (AR), inconsistency rate (IR), and coverage rate (CR) of various VLM methods across multiple dimensions. Overall, Gemini-2.5-Pro-Preview achieves the best or second-best performance in nearly all categories. Notably, it attains the highest AR and CR in most dimensions, such as ``Background'' and  ``Camera Movement'', reaching 92.72\% and 92.09\%, respectively, which demonstrates its strong multimodal understanding capabilities. Meanwhile, its IR remains low, indicating good consistency in its responses. GPT-4o also shows competitive results in certain aspects (such as ``Entity Shape'' and ``Lighting''), but its overall performance is still slightly behind the Gemini series. In contrast, smaller models like LLaVA-Video-7B and Qwen2VL-7B lag significantly in all metrics, especially in complex scenarios and fine-grained attributes (\eg, ``Blur'' and ``Camera Movement''). In summary, as model scale and multimodal capabilities increase, VLMs exhibit substantial improvements in video understanding tasks, with Gemini-2.5-Pro-Preview currently demonstrating the best overall performance.

Fig.~\ref{fig:fig10} and Fig.~\ref{fig:fig11} compare the performance of various Vision-Language Models (VLMs) in terms of caption length, coverage rate (CR), and inconsistency rate (IR). As shown in Figure 1, advanced models such as Gemini-2.5-Pro-Flash, Gemini-2.5-Pro-Preview, and GPT-4o generate much longer captions and achieve higher coverage rates compared to earlier models like Qwen2VL-7B and VideoLLaMA3-7B. Despite the increase in caption length, the inconsistency rate remains low and stable across all models.

Fig.~\ref{fig:fig11} further illustrates the distribution of caption word lengths. Simpler models tend to produce shorter captions, while the more advanced models generate a wider range of longer captions. Overall, these results indicate that recent VLMs are capable of producing more detailed and comprehensive image descriptions without sacrificing consistency.

\subsection{The Impact of Evaluator Selection on the Results}\label{exp_app1}

\begin{table*}[htbp]
\renewcommand{\arraystretch}{0.6}
\caption{The accuracy (AR), inconsistency rate (IR), and coverage rate (CR) the captions of Gemini-2.5-Pro-Preview, GPT-4o and Qwen2.5-VL-72B, where we compare the results of Gemini-2.5-Pro-Preview, GPT-4o, and Manual comprehensive evaluation. The symbol ``$\uparrow$"  indicates that the larger the value, the better; The symbol ``$\downarrow$" indicates that the smaller the value, the better.}
\label{tab:results_3}
\resizebox{\textwidth}{!}{%
\begin{tabular}{@{}l|ccccccccc@{}}
\toprule
 \multirow{3}{*}{Methods}& \multicolumn{3}{c|}{Gemini-2.5-Pro-Preview}& \multicolumn{3}{c|}{GPT-4o}& \multicolumn{3}{c}{Qwen2.5-VL-72B}\\
 \cline{2-10} 
 & AR$\uparrow$ & IR$\downarrow$ &\multicolumn{1}{c|}{CR$\uparrow$}& AR$\uparrow$ & IR$\downarrow$ & \multicolumn{1}{c|}{CR$\uparrow$} & AR$\uparrow$ & IR$\downarrow$ & CR$\uparrow$  \\
 
 \midrule 
Manual &76.87&9.33&\multicolumn{1}{c|}{87.42} &65.47&13.33&\multicolumn{1}{c|}{74.22} &62.76&11.33&70.27\\
GPT-4o &76.73&10.28&\multicolumn{1}{c|}{85.52} &66.49&10.24&\multicolumn{1}{c|}{74.07} &63.08&10.98&70.86  \\ 
Gemini-2.5-Pro-Preview &73.88&14.62&\multicolumn{1}{c|}{86.52} &63.08&15.90&\multicolumn{1}{c|}{75.01} &61.20&16.02&72.87\\

\bottomrule
\end{tabular}%
}
\end{table*}

We randomly selected 100 samples from the dataset and evaluated the quality of video captions generated by different models (\eg, Gemini-2.5-Pro-Preview, GPT-4o and Qwen2.5-VL-72B) using three methods: manual evaluation, method~b, and method~c. The results are shown in Table~\ref{tab:results_3}. As observed, although there are some differences among the automatic evaluation methods, their results are generally consistent with the manual evaluation, and the performance gaps are not significant. This demonstrates that the proposed automatic evaluation process can effectively support the rapid assessment and optimization of video caption generation models.

\subsection{Limitations}\label{exp_app2}

A primary limitation of our proposed evaluation framework is its dependence on the capabilities of the large language model (LLM) employed as the evaluation expert. The use of different LLMs as evaluators may yield variations in the absolute quantitative scores assigned to model outputs. Nevertheless, as long as the same LLM is consistently used for evaluation and possesses sufficiently advanced capabilities (\eg, GPT-4.1 or Gemini 2.5), it can reliably distinguish the relative quality of video captions generated by different models. In general, the reliability of the evaluation results improves with the strength of the LLM serving as the evaluator. We anticipate that, with the ongoing rapid progress in LLM development, the accuracy and robustness of our automatic evaluation framework will continue to improve, further enhancing its utility for video captioning assessment.

\iftoggle{isSubmission}{
  \section*{NeurIPS Paper Checklist}

\begin{enumerate}

\item {\bf Claims}
    \item[] Question: Do the main claims made in the abstract and introduction accurately reflect the paper's contributions and scope?
    \item[] Answer: \answerYes{} % Replace by \answerYes{}, \answerNo{}, or \answerNA{}.
    \item[] Justification: The main claims in the abstract and introduction are consistent with the paper's actual contributions and scope.

\item {\bf Limitations}
    \item[] Question: Does the paper discuss the limitations of the work performed by the authors?
    \item[] Answer: \answerYes{} % Replace by \answerYes{}, \answerNo{}, or \answerNA{}.
    \item[] Justification: The paper includes a dedicated section (A.3 Limitations) discussing the limitations of the work.

\item {\bf Theory assumptions and proofs}
    \item[] Question: For each theoretical result, does the paper provide the full set of assumptions and a complete (and correct) proof?
    \item[] Answer: \answerNo{} % Replace by \answerYes{}, \answerNo{}, or \answerNA{}.
    \item[] Justification: In this paper, we introduce a dataset for assessing the quality of video captions. Notably, our approach does not depend on theoretical assumptions, nor does it require formal theoretical proofs.

    \item {\bf Experimental result reproducibility}
    \item[] Question: Does the paper fully disclose all the information needed to reproduce the main experimental results of the paper to the extent that it affects the main claims and/or conclusions of the paper (regardless of whether the code and data are provided or not)?
    \item[] Answer: \answerYes{} % Replace by \answerYes{}, \answerNo{}, or \answerNA{}.
    \item[] Justification: All data used in this study are made available, including the original videos, the annotated QA-pair files, the video captions generated by the VLMs described in the paper, and the corresponding evaluation results produced by the expert evaluators.

\item {\bf Open access to data and code}
    \item[] Question: Does the paper provide open access to the data and code, with sufficient instructions to faithfully reproduce the main experimental results, as described in supplemental material?
    \item[] Answer: \answerYes{} % Replace by \answerYes{}, \answerNo{}, or \answerNA{}.
    \item[] Justification: All data utilized in this study are made publicly available, including the original videos, annotated QA-pair files, video captions generated by the VLMs discussed in the manuscript, and the corresponding evaluation results provided by expert evaluators. Furthermore, we release the evaluation scripts as well as the prompts employed for video caption generation, to facilitate reproducibility and further research.

\item {\bf Experimental setting/details}
    \item[] Question: Does the paper specify all the training and test details (e.g., data splits, hyperparameters, how they were chosen, type of optimizer, etc.) necessary to understand the results?
    \item[] Answer: \answerYes{} % Replace by \answerYes{}, \answerNo{}, or \answerNA{}.
    \item[] Justification: These details can be found in the experimental section of the paper and in the supplementary materials.

\item {\bf Experiment statistical significance}
    \item[] Question: Does the paper report error bars suitably and correctly defined or other appropriate information about the statistical significance of the experiments?
    \item[] Answer: \answerYes{} % Replace by \answerYes{}, \answerNo{}, or \answerNA{}.
    \item[] Justification: To analyze the variability and potential sources of error in the evaluation results, we compared the assessments provided by different expert evaluators.

\item {\bf Experiments compute resources}
    \item[] Question: For each experiment, does the paper provide sufficient information on the computer resources (type of compute workers, memory, time of execution) needed to reproduce the experiments?
    \item[] Answer: \answerNo{} % Replace by \answerYes{}, \answerNo{}, or \answerNA{}.
    \item[] Justification: As our primary contribution is to provide a benchmark dataset for video caption quality assessment, information about computational resources (\eg, worker type, memory, execution time) is not the focus of our work, since we do not aim to optimize model efficiency. Our evaluations are performed using the currently available VLMs, and details regarding resource usage can be found in the respective technical reports of these models.
    
\item {\bf Code of ethics}
    \item[] Question: Does the research conducted in the paper conform, in every respect, with the NeurIPS Code of Ethics \url{https://neurips.cc/public/EthicsGuidelines}?
    \item[] Answer: \answerYes{} % Replace by \answerYes{}, \answerNo{}, or \answerNA{}.
    \item[] Justification: the research conducted in the paper fully conforms to the NeurIPS Code of Ethics in every respect.

\item {\bf Broader impacts}
    \item[] Question: Does the paper discuss both potential positive societal impacts and negative societal impacts of the work performed?
    \item[] Answer: \answerYes{} % Replace by \answerYes{}, \answerNo{}, or \answerNA{}.
    \item[] Justification: This issue has been discussed in the experimental section of the paper.
    
\item {\bf Safeguards}
    \item[] Question: Does the paper describe safeguards that have been put in place for responsible release of data or models that have a high risk for misuse (e.g., pretrained language models, image generators, or scraped datasets)?
    \item[] Answer: \answerNo{} % Replace by \answerYes{}, \answerNo{}, or \answerNA{}.
    \item[] Justification: This paper does not involve the release of any data or models that pose a high risk for misuse, such as pretrained language models, image generators, or scraped datasets.

\item {\bf Licenses for existing assets}
    \item[] Question: Are the creators or original owners of assets (e.g., code, data, models), used in the paper, properly credited and are the license and terms of use explicitly mentioned and properly respected?
    \item[] Answer: \answerYes{} % Replace by \answerYes{}, \answerNo{}, or \answerNA{}.
    \item[] Justification: All assets used in this paper are properly credited. The data used in our experiments are obtained from publicly available open-source datasets, and we have explicitly cited the original sources in the paper. We have carefully reviewed and respected the licenses and terms of use associated with these datasets.

\item {\bf New assets}
    \item[] Question: Are new assets introduced in the paper well documented and is the documentation provided alongside the assets?
    \item[] Answer: \answerYes{} % Replace by \answerYes{}, \answerNo{}, or \answerNA{}.
    \item[] Justification: This paper presents a new dataset, which is openly accessible via Hugging Face and GitHub. We have included thorough documentation and detailed descriptions to ensure transparency and ease of use for other researchers.

\item {\bf Crowdsourcing and research with human subjects}
    \item[] Question: For crowdsourcing experiments and research with human subjects, does the paper include the full text of instructions given to participants and screenshots, if applicable, as well as details about compensation (if any)? 
    \item[] Answer: \answerNo{} % Replace by \answerYes{}, \answerNo{}, or \answerNA{}.
    \item[] Justification: this paper does not involve any crowdsourcing experiments or research with human subjects.

\item {\bf Institutional review board (IRB) approvals or equivalent for research with human subjects}
    \item[] Question: Does the paper describe potential risks incurred by study participants, whether such risks were disclosed to the subjects, and whether Institutional Review Board (IRB) approvals (or an equivalent approval/review based on the requirements of your country or institution) were obtained?
    \item[] Answer: \answerNo{} % Replace by \answerYes{}, \answerNo{}, or \answerNA{}.
    \item[] Justification: This paper does not involve research with human subjects, and therefore IRB approval or equivalent review was not required.

\item {\bf Declaration of LLM usage}
    \item[] Question: Does the paper describe the usage of LLMs if it is an important, original, or non-standard component of the core methods in this research? Note that if the LLM is used only for writing, editing, or formatting purposes and does not impact the core methodology, scientific rigorousness, or originality of the research, declaration is not required.
    %this research? 
    \item[] Answer: \answerYes{} % Replace by \answerYes{}, \answerNo{}, or \answerNA{}.
    \item[] Justification: The paper details the use of LLMs as a key and original part of the research methodology. In particular, LLMs are utilized as evaluation experts and actively participate in the process of generating video captions.

\end{enumerate}
}

\end{document}